\documentclass[fleqn]{article}

\usepackage{PRIMEarxiv}
\usepackage{bibspacing}
\usepackage[utf8]{inputenc} 
\usepackage[T1]{fontenc}    
\usepackage{appendix}
\usepackage{hyperref}       
\usepackage{url}            
\usepackage{booktabs}       
\usepackage{amsfonts}       
\usepackage{nicefrac}       
\usepackage{microtype}      
\usepackage{lipsum}
\usepackage{fancyhdr}       
\usepackage{graphicx} 
\usepackage{verbatim} 
\graphicspath{{media/}}     
\usepackage[notocbib]{apacite}
\usepackage{babel} 
\usepackage{amsmath}
\usepackage{amssymb}

\usepackage{pgffor}
\usepackage{multirow}
\usepackage{float}
\usepackage{array}
\usepackage{mdsymbol}
\usepackage{mathtools}
\usepackage{pdflscape}
\usepackage{float}
\usepackage{enumitem}
\usepackage{MnSymbol}

\usepackage{tabularx}
\usepackage{caption}
\usepackage{subcaption}
\usepackage{booktabs}
\usepackage{adjustbox}
\usepackage{subcaption} 

\usepackage{rotating}
\newcommand{\ip}[2]{\parbox{#1}{\centering\vspace{0.25cm} #2}}

\newcommand{\cm}{$\checkmark$}

\title{Energy Price Modelling: A Comparative Evaluation of four Generations of Forecasting Methods
}

\author{
Alexandru-Victor ANDREI\thanks{\scriptsize  Bucharest University of Economic Studies, Romania. \url{andrei1victor23@stud.ase.ro}}\and 
Georg VELEV \thanks{\scriptsize  Humboldt University of Berlin, Germany.} \and
Filip-Mihai TOMA \thanks{\scriptsize  Bucharest University of Economic Studies, Romania; California Institute of Technology, USA} \and
Daniel Traian PELE \thanks{\scriptsize  Bucharest University of Economic Studies, Romania.
Institute for Economic Forecasting, Romanian Academy.} \and
Stefan LESSMANN \thanks{\scriptsize  Humboldt University of Berlin, Germany. Bucharest University of Economic Studies, Romania.}
}

\usepackage{acro}
\DeclareAcronym{ER}{
  short = ER,
  long = Erdös-Rényi
}

\DeclareAcronym{CSL}{
  short = CSL,
  long = causal structure learning
}

\DeclareAcronym{SF}{
  short = SF,
  long = scale-free
}

\DeclareAcronym{LLM}{
  short = LLM,
  long = large language model
}

\begin{document}
\setlength{\bibitemsep}{0.9\baselineskip plus 0.05\baselineskip minus .05\baselineskip}

\maketitle

\begin{abstract}
Energy is a critical driver of modern economic systems. Accurate energy price forecasting plays an important role in supporting decision-making at various levels, from operational purchasing decisions at individual business organizations to policy-making. A significant body of literature has looked into energy price forecasting, investigating a wide range of methods to improve accuracy and inform these critical decisions. Given the evolving landscape of forecasting techniques, the literature lacks a thorough empirical comparison that systematically contrasts these methods.

This paper provides an in-depth review of the evolution of forecasting modeling frameworks, from well-established econometric models to machine learning methods, early sequence learners such LSTMs, and more recent advancements in deep learning with transformer networks, which represent the cutting edge in forecasting. We offer a detailed review of the related literature and categorize forecasting methodologies into four model families. We also explore emerging concepts like pre-training and transfer learning, which have transformed the analysis of unstructured data and hold significant promise for time series forecasting. We address a gap in the literature by performing a comprehensive empirical analysis on these four family models, using data from the EU energy markets, we conduct a large-scale empirical study, which contrasts the forecasting accuracy of different approaches, focusing especially on alternative propositions for time series transformers.
\end{abstract}


\section{Introduction}

Energy plays a key role in policy-making, operational efficiency, and strategic planning across all sectors in any economy. Accurate energy price forecasting is essential for minimizing costs, optimizing resource allocation, and reducing risks for stakeholders in the energy supply chain. Within energy forecasting, the ability to accurately predict energy prices is particularly critical due to the volatility and complexity of energy markets, which warrants the need for improved forecasting accuracy and real-time data evaluation.

While forecasting is fundamental across various domains, there is a notable gap in energy price forecasting. Most energy forecasting studies have focused on more predictable variables, such as energy generation, consumption patterns, and supply dynamics, which tend to exhibit more stable trends. In contrast, the volatile nature of energy prices presents greater forecasting challenges. A comprehensive review of the  literature shows a scarcity of research on energy price forecasting (for details, see Table 1). This may seem surprising given the importance of price forecasting in the energy sector and while current literature has covered energy forecasting using different methodologies (Table 1), there still is a gap in comparing performances of different model generations over the past years, that could offer a more comprehensive view of  energy price forecasting methods. As such, the need to benchmark and evaluate state-of-the-art forecasting methodologies in energy price modeling becomes critical. This is not only important for research, but has direct applicability for policymakers, energy stakeholders, and consumers that rely on accurate price forecasts for adequate decision-making.

Different methodologies that use time-series in analyzing trends and generating forecasts have been developed. Such methods range from classical econometric models to complex machine learning techniques, and more recently, to advanced deep learning frameworks. Particularly, transformer networks, a subset of deep learning models, have had a significant influence in many branches of data analysis, including time series forecasting. These transformer models offer encouraging improvements over classical approaches due to their capacity for handling long-term dependencies and very complex patterns. 

However, despite the potential of these advanced techniques, there is a need to evaluate and benchmark their performance in the domain of energy price forecasting. Without a clear comparative analysis, the relative strengths, limitations, and practical applicability of these methods remain unclear, highlighting the importance of systematic benchmarking to establish a robust foundation for effective energy price forecasting. Additionally, our literature review reveals another significant gap: to date, no Large Language Model (LLM) has been used in energy sector time series forecasting. Thus, this presents an opportunity to explore whether these models, building on their demonstrated success in handling complex patterns in natural language, can be adapted to improve forecasting accuracy.

This paper aims to address these gaps by conducting a thorough review on multiple generations of forecasting methodologies. It categorizes the different approaches and presents a comprehensive benchmarking experiment using data from the European energy markets, more specifically - the electricity markets. This experiment evaluates the forecasting accuracy of classical econometric methods, machine learning architectures, sequence models, the latest deep transformer networks, and for the first time, Large Language Models. In addition, the paper explores novel concepts such as pre-training and transfer learning to reveal their applicability in time series forecasting.

In this study, our focus is specifically on electricity price forecasting, a critical subset of energy forecasting. While energy forecasting encompasses a broad range of sectors—including oil, gas, and renewables—electricity presents unique challenges and dynamics. Due to its non-storable nature, electricity demand and supply must be balanced in real time, leading to heightened price volatility. Additionally, electricity prices are influenced by both generation sources, such as renewables, and demand factors that vary throughout the day and across seasons. By concentrating on electricity price forecasting within the broader context of energy, our study aims to capture the complexities inherent in one of the most volatile and data-intensive energy sectors, providing insights that are relevant for stakeholders in the energy supply chain, policy, and market operations

Through a comparative analysis, the study seeks to answer the following key questions: How do state-of-the-art forecasting methods perform relative to their predecessors? Can LLM-based models deliver significant improvements in forecasting accuracy?

By addressing these questions, this research aims to advance efforts to improve energy price forecasting accuracy, thereby facilitating more informed decision-making in policy development and energy management.\\

\section{Related literature: Energy Time Series Forecasting}

\subsection{Classification of Time Series Forecasting Techniques}
In this section, we present a taxonomy of existing time series forecasting models. The classification into different families is based on: their  methodological innovations, their ability to capture complex patterns (expressive power) and the computational resources necessary to train the models. We motivate our classification as follows: (1) to systematically capture the evolution of forecasting methodologies, highlighting the key innovations and technical shifts that have advanced the field; (2) to compare models based on their complexity, computational requirements, and ability to capture intricate time-series patterns, facilitating a structured analysis of each generation’s strengths and limitations; and (3) to underscore the growing adaptability and applicability of forecasting methods, from foundational econometric models to cutting-edge LLMs, in addressing increasingly complex forecasting needs in the energy sector.\\ 

\begin{figure}[h]
\centering
\includegraphics[scale=0.35]{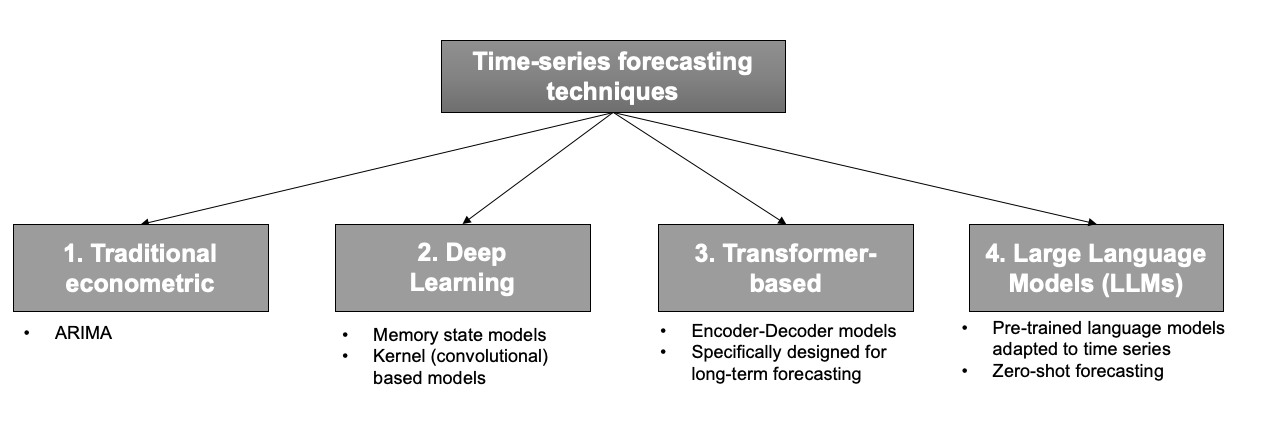}
\caption{Taxonomy of existing families (generations) of time series forecasting methods}
\label{fig:FourGenerationsForecastingMethods}
\end{figure}

Figure \ref{fig:FourGenerationsForecastingMethods} shows the classification of four distinct generations of forecasting models.
The \textbf{first} generation includes traditional econometric models, which overall laid the foundation for time series analysis. These models employ relatively simple, linear techniques and generally require fewer computational resources for training compared to the more advanced methods introduced in subsequent generations. A well-known example of a first-generation time series model is ARIMA, well-suited for modeling linear trends based on univariate time series data. The \textbf{second} generation of techniques consists of neural-based time series models, which are capable of modeling nonlinear dependencies between multivariate inputs and target time series. Compared to first generation models, these exhibit both higher algorithmic complexity and enhanced predictive power, thanks to the use of nonlinear operations such as recurrent and convolutional layers, which can capture more complex time-series patterns as compared to linear models. Additionally, neural-based models can be configured to any depth or width, allowing for a flexible number of stacked nonlinear operations, vertically or horizontally. For this reason, we associate the second generation with deep learning techniques, whose training generally demands more computational resources than linear models. The \textbf{third} generation consists of newly developed models for time series forecasting, that use transformer-based frameworks. As with models from the second generation, transforms are also neural networks: however, they leverage unique attention-based mechanisms to capture long-range dependencies in temporal data more effectively. Additionally, most transformer models are specifically tailored for long-term forecasting, which enhances their expressive power as compared to models from the first two generations. The \textbf{fourth} generation is comprised of the latest advancements in time series forecasting models, drawing inspiration from Large Language Models (LLMs). This family of techniques includes an initial pre-training phase, which leverages a large corpus of time series. Although pre-training requires significantly more computational resources compared to training task-specific models from earlier generations, deploying pre-trained models reduces computational demands due to their zero-shot forecasting capabilities. This pre-training phase eliminates the need to train models from the fourth generation on a specific dataset. Unlike second and third generation models, these foundation models typically work with univariate input time series.\\

\subsection{Related studies in energy and electricity price forecasting}
The field of energy forecasting has witnessed considerable methodological advancements in recent years, driven by increasingly complex energy systems and the advent of advanced computational techniques. Additionally, a growing trend in energy price forecasting research is the use of public datasets for enhanced reproducibility. However, some results remain difficult to verify due to certain data confidentiality. Geographically diverse datasets, primarily from the United States, China, and several European nations, include a wide range of variables such as electricity consumption, load, renewable energy generation and battery life expectancy. Temporal resolutions in these datasets vary widely, ranging from seconds to weeks, depending on the study.

To date, benchmark studies on energy price forecasting have primarily compared methods from the second and third generations. For example, \shortcite{11_Sherozbek} and \shortcite{13_HuangJH} explored transformer-based models for forecasting photovoltaic (PV) power output and building energy consumption, while \shortcite{14_Ramos} compared various deep learning models for forecasting residential energy consumption. Similarly, \shortcite{17_Jiang} tested hybrid methods from the 2nd and 3rd generations for wind speed forecasting, while \shortcite{56_Oh} compared various neural network architectures across different energy-related datasets, yielding mixed results depending on the used dataset. Additionally, \shortcite{62_Zhao} provided a comprehensive review of deep learning techniques for battery health prognostics. Existing reviews of machine learning methods of energy price forecasting up to 2022 (e.g., (\shortcite{WERON2014}; \shortcite{NOWOTARSKI2018}; \shortcite{GHODDUSI2019}; \shortcite{HONG2020}; \shortcite{LAGO2021}; \shortcite{Jedrzejewski2022}) omit transformer architectures and outline several guidelines for research in electricity price forecasting. These include ensuring reproducibility of results, through open-source code and accessible datasets where possible, and the need for robust model error measurement procedures. Additionally, these studies also emphasize the importance of conducting significance tests to evaluate a model's performance relative to others.

Deregulation in electricity markets has introduced substantial price risks for market participants—risks that differ significantly from those observed in other commodity or financial markets \cite{seifert2007modelling}. As a non-storable commodity, electricity requires real-time balancing of supply and demand, contributing to its distinctive price behavior \cite{bierbrauer2007spot}. This leads to unique characteristics in electricity spot prices, including multiple seasonal patterns, a strong mean-reverting tendency, persistent volatility, frequent price jumps, short-lived spikes, and an inverse leverage effect. Together, these factors make accurately modeling electricity prices a complex challenge.

In response to these complexities, recent studies have leveraged machine learning models to improve electricity price forecasting. For example, \cite{tschora2022electricity} demonstrate that incorporating previously unused predictive features, such as neighboring countries' price histories, can significantly enhance forecast accuracy across the European market. \cite{gabrielli2022data} further highlight the potential of data-driven approaches through a Fourier analysis model that effectively captures both base trends and high-volatility components in electricity prices. Their model generalizes well to unseen electricity markets, underscoring the value of finely-resolved, hybrid data-driven approaches for long-term forecasting, especially in renewable energy investment contexts.

One recent study compares forecasting methodologies, including LLMs, and finds that while TimeGPT performs well in stable market conditions, models like KAN and PatchTST are better suited to complex, volatile scenarios, highlighting both the strengths and limitations of LLMs for time series forecasting and the need for further empirical testing on other asset types \shortcite{alonso2024large}. However, to the best of our knowledge, no studies have employed LLMs for electricity price forecasting. We also note other gaps in energy price forecasting, namely: the limited use of transformer models, the absence of research employing LLMs, and the use of walk-forward validation techniques. In addition, studies that focus on electricity prices are limited to single country analyses, such as Spain, Greece, and Germany or on specific regions like the Germany-Luxembourg region, such as those from \shortcite{2_Zhang}, \shortcite{3_Xu}, \shortcite{8_Zhong}, and \shortcite{37_Cantillo-Luna}. In contrast, our study uses a comprehensive dataset of electricity prices covering 27 European countries, which ensures a high degree of homogeneity in regulation and price formation for the EU countries, while still minimizing the risk of data discrepancies across the broader European dataset, providing a strong foundation for forecasting.



\subsection{Time series models for forecasting}
In this section, we provide a structured overview of time series models representative of the four generations of forecasting techniques that allow us to underscore research gaps in electricity price forecasting. Additionally, we point out the selection criteria for including a specific set of approaches in the subsequent empirical analysis.\\ 

\begin{table}
\centering
\resizebox{\textwidth}{!}{
\begin{tabular}{|c|c|c|c|c|c|c|c|c|c|c|c|}
				\hline
				\multirow{3}{*}{\ip{2.5cm}{\textbf{Time Series Model (Reference)}}} &\multicolumn{6}{c|}{\textbf{Forcasting Approach}}  & \multicolumn{2}{c|}{\textbf{Methodology}} & \multicolumn{2}{c|}{\textbf{Dataset}} & \multirow{3}{*}{\ip{3cm}{\textbf{Walk-Foward Performance Validation}}} \\
				\cline{2-11}
				& \multirow{2}{*}{\textbf{Econometric}} & \multicolumn{3}{c|}{\textbf{Deep Learning}} & \multirow{2}{*}{\textbf{Transsformer}}& \multirow{2}{*}{\textbf{LLM}}& \multirow{2}{*}{\textbf{Decomposition}}& \multirow{2}{*}{\ip{2cm}{\textbf{Novel Self-Attention Module}}}& \multirow{2}{*}{\ip{2cm}{\textbf{EU Electricity Prices}}} & \multirow{2}{*}{\ip{2cm}{\textbf{Snythetic Dataset}}}& \\
				\cline{3-5}
				&& \ip{1cm}{\textbf{MLP based}} & \ip{1cm}{\textbf{CNN based}} & \ip{1.5cm}{\textbf{Recurrent based}} &&&&&&&\\
				\hline
				DLinear & & & & & & & & & & &\\
                \shortcite{nlinear_dlinear}&-&\cm&-&-&\cm&-&\cm&-&-&-&-\\
				\hline
                 NLinear & & & & & & & & & & &\\
				\shortcite{nlinear_dlinear}&-&\cm&-&-&\cm&-&-&-&-&-&-\\
				\hline
                TSMixer & & & & & & & & & & &\\
				\shortcite{TSMixer}&\cm&\cm&-&-&\cm&-&-&-&-&-&-\\
				\hline
                Autoformer & & & & & & & & & & &\\
				\shortcite{Autoformer}&\cm&-&-&-&\cm&-&\cm&\cm&-&-&-\\
				\hline
                Basisformer & & & & & & & & & & &\\
				\shortcite{Basisformer}&-&\cm&\cm&-&\cm&-&-&\cm&-&-&-\\
				\hline
                Informer & & & & & & & & & & &\\
				\shortcite{Informer}&\cm&-&-&\cm&-&-&-&\cm&-&-&-\\
				\hline
                PatchTST & & & & & & & & & & &\\
				\shortcite{PatchTST}&-&-&\cm&-&\cm&-&-&-&-&-&-\\
				\hline
                Quatformer & & & & & & & & & & &\\
				\shortcite{Quatformer}&-&-&\cm&\cm&\cm&-&\cm&\cm&-&-&-\\
				\hline
                Chronos & & & & & & & & & & &\\
				\shortcite{ansari_chronos_2024}&\cm&\cm&-&-&\cm&\cm&-&-&-&\cm&-\\
				\hline
                TimesFM & & & & & & & & & & &\\
				\shortcite{das_decoder-only_2024}&-&\cm&-&-&\cm&\cm&-&-&-&\cm&-\\
				\hline
\end{tabular}
		}
\caption{Overview of the second, third and fourth generations of time series forecasting techniques}
\label{tab:fourgenerations}
\end{table}

Table \ref{tab:fourgenerations} classifies models by forecasting approach, indicating the family each model belongs to and the benchmark models included in each corresponding study. Although most studies compare recent techniques to transformer-based models, the varied selection of baseline models across studies highlights the need for a comprehensive benchmarking effort that spans all four generations. 
Furthermore, Table \ref{tab:fourgenerations} reveals a notable gap in the adoption of walk-forward validation for long-term time series forecasting, despite its critical role in testing model robustness over time. Since the evaluation of fairly complex transformers-based models on well-known test subsets could potentially result into issues related to overfitting, the application of the walk-forward validation approach would shed light on the robustness of the performance of the models over time. Furthermore, the LLM models - Chronos and TimesFM - are  evaluated only on synthetic time series data, not on empirical electricity prices. The former can be simulated with varying time-dependent components, e.g., upward and downward trends, multiple distinct periodic patterns, etc., in order to explore the performance of recently introduced models in long-term forecasting scenarios with varying complexity. Additionally, Table \ref{tab:fourgenerations} shows none of the models from the second, third and fourth generation have been tested against (EU) electricity price time series. Our empirical research addresses this by focusing on long-term electricity price forecasting for 27 EU countries, aproviding a unified regulatory and pricing framework that minimizes data inconsistencies, as discussed in \ref{subsec:energydata}. The remainder of this section details the methodologies employed by the long-term time series forecasting techniques in Table \ref{tab:fourgenerations}. While presenting the forecasting methodology, we highlight the selection criteria for including the models from Table \ref{tab:fourgenerations} in our empirical analysis. \\

\textbf{First generation: conventional linear models}\\
Table \ref{tab:fourgenerations} shows that four out of ten studies presenting more recent techniques include econometric benchmarks, with ARIMA being a common choice.
ARIMA (Autoregressive Integrated Moving Average) models, known for their simplicity and interpretability, decompose time series into three components: autoregressive (AR), integration (I), and moving average (MA). This approach enables ARIMA to effectively model linear trends, making it a foundational technique in time series analysis. Given its enduring role in time series research and its documented effectiveness for univariate forecasting, we include ARIMA as a benchmark in our empirical study.
ARIMA has a long standing position in the literature of time series analysis, as it has been extensively used and is well-documented. \shortcite{Hyndman2018} provide a comprehensive overview of various forecasting methods, including ARIMA. Particularly, \shortcite{24_Santos} and \shortcite{2_Zhang} use ARIMA models in benchmarking studies with electricity price data. Therefore, we include ARIMA in the set of benchmark models of our empirical study.\\

\textbf{Second generation: deep learning}\\
The \textbf{second generation} of models marks a significant leap in forecasting methodology, with the introduction of neural networks that expand the capacity of time series models to capture nonlinear relationships and intricate temporal dependencies. These models, powered by deep learning architectures, set the stage for improved accuracy in multivariate forecasting tasks, a critical advancement in fields like energy forecasting, where data often exhibits complex, interconnected patterns. Below, we discuss prominent second-generation models that illustrate these advancements and address some of the key gaps in traditional forecasting methods.

\shortciteA{TSMixer} introduce TSMixer, an architecture specifically  adapted for time series forecasting. The model uses an all-MLP (multi-layer perceptron) setup that leverages cross-variate information while retaining the ability to capture temporal patterns. TSMixer excels with its capability to extract information by mixing operations along time and features dimensions. Thus, it is very useful in multivariate forecasting activities where cross-variate information might not have a positive contribution. TSMixer thus addresses one of the key limitations of traditional models: the risk of overlooking the intricate interplay between multiple time series variables. With a performance comparable to univariate models on standard benchmarks, TSMixer achieves state-of-the-art results in industrial applications, highlighting its adaptability to real-world scenarios requiring cross-variate insights. Recently, \cite{meng2024enhancing} find apply deep learning models to improve RMB/USD exchange rate predictions, overcoming the limitations of traditional models in handling complex, non-linear data. They find that among LSTM, CNN, and transformer-based architectures, TSMixer is the most effective, outperforming both LSTM and CNN.This finding supports our decision to exclude CNN and LSTM frameworks from our empirical approach, as TSMixer demonstrated superior performance in handling the complexities of exchange rate prediction.

NLinear and DLinear are direct applications of LTSF\footnote{long-term time series forecasting}-Linear models introduced by \shortcite{nlinear_dlinear}. The two models show major advances in in decomposition and normalization methods for long-term time series forecasting. DLinear uses linear layers in a decomposition scheme inspired by FEDformer and Autoformer, where time series data is split into trend and seasonal components, with the trend component processed using a moving average kernel. Then, these two components are processed by two one-layer networks for direct multi-step forecasting. This decomposition method handles data trends and improves the performance of a vanilla linear model in scenarios with evident trend. In contrast, NLinear applies a normalization method to deal with distribution shifts in datasets. Firstly, the model removes from the input the last value from the sequence before passing it through a linear layer. Then, after the linear transformation, the removed value is restored to the input before the final prediction.

Long Short-Term Memory networks (LSTMs) are a type of recurrent neural networks (RNN). LSTMs are broadly used for sequence prediction tasks such as speech recognition, image analysis, and time series forecasting. With features such as local connectivity, translation invariance, and hierarchical structures, LSTMs are well suited at capturing long-term dependencies in temporal data, addressing the limitations of traditional methods that struggle with extended sequences. Their architecture allow LSTMs to model complex temporal patterns and dependencies, making them suitable for a broad range of applications.

Convolutional Neural Networks (CNNs), initially designed for image recognition tasks, such as detection, segmentation, and image classification, have also been proven effective in time series forecasting. They automatically extract features, bypassing the classical machine learning approach of manual feature engineering. CNNs have also achieved impressive results in domains beyond image processing, such as speech recognition, natural language processing, and time series analysis. In time series forecasting, CNNs offer a robust mechanism to capture localized patterns, further contributing to the advancement of multivariate forecasting capabilities within the second generation of models.
\\

\textbf{Third generation: transformers}\\
Transformer models, initially developed with an encoder-decoder architecture for natural language processing, have been widely adapted for time series forecasting due to their ability to capture long-range dependencies through attention mechanisms. This third generation of models represents a notable advancement in time series forecasting by introducing architectures tailored to complex temporal dependencies and long forecasting horizons—critical in fields like energy price forecasting.

\shortciteA{Autoformer} introduce Autoformer, which represents a decomposition-based transformer architecture specifically designed for long forecasting horizons. The deep decomposition modules of Autoformer account for different time dependent patterns occurring in complex time series, e.g., upward and downward trends,  plateau intervals, fluctuations, steep drops and rises etc. Since in our empirical research we are concerned with forecasting electricity price data, which can exhibit varying nonlinear temporal components, we include Autoformer in our set of benchmark models. The decoder's autocorrelation modules of Autoformer are applied to the latent seasonal components in order to measure period-based similarities among the time-series. Autoformer generates predictions for both temporal components of the prediction sequence, i.e., seasonal and trend-cyclical components, the sum of which produces the final forecasts. 

Building on the decomposition approach, \shortciteA{Quatformer} developed Quatformer, which also models periodic and trend-related patterns but introduces a novel learning-to-rotate attention mechanism. This mechanism which allows for the representation extraction of variable periods and phase shifts. Therefore, we include Quatformer in our benchmark study in order to examine whether Quatformer's algorithmic innovation can produce more accurate forecasts than Autoformer's computations. \shortciteA{FEDformer} present FEDformer's architecture, which bears a lot of similarities with Autoformer, as FEDformer also uses trend-seasonal decomposition blocks as well as self-attention and cross-attention Fourier Transformation-based blocks. For this reason, we refrain from including FEDformer in our empirical study.\\

Informer, on the other hand, adopts a unique architecture that bypasses decomposition, favoring efficiency over complex seasonal-trend representations \shortcite{Informer}. In contrast to Autoformer, FEDformer and Quatformer, Informer's architecture does not incorporate decomposition modules and it is more efficient than most transformer-based models. For instance, Informer downsamples the latent representation of the input time series by applying information distilling operations, which are realized with convolutional layers followed with max pooling modules. Additionally, Informer generates forecasts in one forward operations instead of applying the computationally expensive dynamic (i.e., autoregressive), decoding of the output sequences. Due to its efficient computations, we include Informer in the set of baseline techniques of our empirical analysis. While Basisformer adopts Autoformer’s idea for learnable basis vectors, namely latent trend and seasonal components extracted from the input time series, Basisformer models the basis in a self-supervised manner by applying contrastive learning to the historical view of the time series-the input sequences, and the future view of the data, namely the prediction sequences \shortcite{Basisformer}. The reason for modeling the contrastive views is to select basis representations, which are consistent across both views. Since self-supervised learning can potentially enrich the modeling of latent basis vectors, which can in turn positively affect the expressive power of the model, we include Basisformer in our empirical analysis. 

PatchTST introduced by \shortcite{PatchTST} facilitates the modeling of subseries-level representations or “patches,” which capture localized temporal patterns across timesteps. In this regard, patching improves the memory consumption by reducing the number of input tokens to PatchTST's encoder by a factor of the stride. Furthermore, PatchTST utilizes channel-independence by splitting the multivariate time-series into independent channels, and extracting latent representations for each univariate time series. Since the forward pass uses shared weights for each univariate input, the channel-independence mechanism enables memory efficient learning across all channels in the original multivariate time-series. Therefore, we include PatchTST in our research due to advantages related to scalability. 

Lastly, Crossformer, unlike other transformers, adopts Dimension-Segment-Wise embedding to capture cross-feature relationships alongside temporal dependencies \shortcite{Crossformer}. This makes the application of a Two-Stage Attention modules necessary, in order to capture the correlational structure among the time steps as well as among the time series attributes. Since the Two-Stage Attention computation is highly likely to have a negative impact on Crossformer's training time, we refrain from including it in our benchmark study.\\


\textbf{Fourth generation: Large Language Models}\\
LLM-based models have emerged as a promising alternative for time-series forecasting, including in the energy price sector, offering distinct advantages over traditional Transformers and other deep learning models. While Transformers and deep learning models effectively capture temporal dependencies and complex patterns in time series data, LLM-based models leverage extensive pre-training on vast text corpora, enabling them to incorporate broader contextual knowledge and nuances. Recently, LLM-based models have been shown to act like general pattern learning machines, rather than being language specific, making them ideal for time-series forecasting \shortcite{mirchandani2023largelanguagemodelsgeneral}. Additionally, LLM-based models exhibit zero-shot forecasting abilities, allowing them to make accurate predictions without extensive task-specific training. This is particularly beneficial in fields such as energy markets, where rapid incorporation of new information into data modelling is important. The zero-shot capabilities stem from the models' ability to generalize knowledge from diverse training data, making them adept at understanding and predicting outcomes based on new, unseen data. This significantly reduces the time and computational resources required for training, making LLM-based models a more efficient and versatile choice for financial data forecasting, including energy prices. \shortcite{nie_survey_2024}. This adaptability, coupled with a general pattern-learning capability, positions LLMs as highly efficient and versatile tools for financial data forecasting, including energy prices.

Building on these strengths, recent research has highlighted the effectiveness of LLMs for forecasting applications. For instance, \shortcite{gruver_large_nodate} demonstrated that LLM-based approaches are superior zero-shot forecasters in time series forecasting. Similarly, \shortcite{zhou_one_2023} develop a pre-trained LM and show that pre-trained models on natural language or images can deliver state-of-the art performance across key time series analysis tasks. Furthermore, superior forecasting performance has also been found by \shortcite{jin_time-llm_2024} in time series with clear trends and patterns, even if not so much in time series that lag periodicity. Recent studies use cross-modality alignments in LLM-based zero-shot forecasting tasks to address the computational costs of these methods \shortcite{liu_timecma_2024}. 

Expanding beyond energy prices, other applications of LLMs demonstrate their versatility in integrating macroeconomic indicators and textual data, where they outperform traditional models in forecasting economic variables. \shortcite{gueta_can_2024}. 

Recent advancements in time series analysis have seen a significant transition towards using more sophisticated language-based models, as comprehensively discussed by Zhang et al. \shortcite{zhang2024largelanguagemodelstime}. For example, Lag-Llama represents a recent foundation model for univariate probabilistic time series forecasting, built on a decoder-only transformer architecture, pre-trained on a diverse corpus of time series data from multiple domains, showcasing strong zero-shot generalization capabilities and state-of-the-art performance when fine-tuned on small, unseen datasets \shortcite{rasul_lag-llama_2024-1}

Building on these foundations, we focus on two cutting-edge approaches on forecasting time series data: Chronos and TimesFM. These models offer new capabilities for analyzing and forecasting complex temporal data, which is particularly relevant for the energy sector and beyond.

Chronos presents a framework for pre-trained probabilistic time series models that uses transformer-based language architectures \shortcite{ansari_chronos_2024}. These transformer architectures, initially developed for processing human language, have been adapted to handle numerical time series data. By converting time series data into a format similar to language, Chronos can apply powerful text analysis tools to such data. This innovative approach enables Chronos to generate multiple possible future scenarios based on historical data, providing a range of probabilistic forecasts. Trained on a diverse set of real and artificially created datasets, Chronos models have shown superior performance in extensive tests. Notably, they demonstrate strong performance on new, unseen datasets, revealing their ability to adapt to various energy market conditions without requiring extensive retraining.

TimesFM, developed by Google Research, introduces a specialized Foundation Model designed specifically for time-series data analysis \shortcite{das_decoder-only_2024}. Inspired by recent breakthroughs in natural language processing, TimesFM excels in forecasting single variable time series with great flexibility. It can handle long sequences of historical data and make predictions over various time horizons, incorporating factors like seasonal patterns. This makes TimesFM particularly useful for energy sector applications, where long-term trends and cyclical patterns are inherently present in the data. The model's versatility extends to incorporating external factors into its forecasts, allowing it to adapt to a wide range of energy market scenarios and prediction tasks.

Unlike the multivariate approaches typical of second and third generation models, TimesFM and Chronos focus exclusively on univariate forecasting. These two fourth generation models are included in this benchmark for several reasons. Firstly, the performance they show highlights their  ability in employing large-scale foundational models for time series forecasting even when limited to single time series data (i.e. for the purposes of our study, single-country predictions). Secondly, the multivariate counterparts to TimesFM and Chronos have the advantage of cross-country data which hints that univariate forecasting from FM models can achieve high levels of accuracy when combined with advanced learning strategies derived from LLMs. This suggests that FM models might compensate for their lack of multivariate data by deeper contextual understanding. All these reasons justify including TimesFM and Chronos in the proposed benchmark study and underscore the advancing capabilities of fourth-generation forecasting models in time series analysis.

By incorporating these state-of-the-art models, this work seeks to expand an empirical overview on time series analysis, exploring their potential to on forecasting accuracy, pattern recognition, and temporal relationships, compared to earlier LLM applications in this domain. Our study makes several notable contributions to the field of time series forecasting, particularly in the context of price prediction, known for complexity due to high volatility and sensitivity to external factors. Firstly, we extend existing benchmarks by focusing specifically on electricity price forecasting, providing insights relevant to financial and economic contexts.  Secondly, our research encompasses a comprehensive evaluation of recently introduced transformer-based models, addressing a gap in prior studies which have not fully explored such architectures for price forecasting. This ensures that our benchmarking experiment is up-to-date with the latest developments in deep learning for time series analysis. Lastly, we incorporate the most recent developments in the field of Large Language Models (LLMs) applied to time series forecasting, such as Chronos and TimesFM. By leveraging these models, we aim to identify patterns and relationships that may not be detectable with traditional methods. This approach not only contributes to the understanding of price forecasting methodologies but also provides a robust empirical benchmark for future research in energy price forecasting.

\section{Data and results}
\subsection{Energy price time series}
\label{subsec:energydata}
In this section, we describe the data retrieval process, as well as providing descriptive statistics of our energy time series dataset.

We obtain hourly wholesale electricity price time series of a total of 27\footnote{Austria, Belgium, Bulgaria, Croatia, Czechia, Denmark, Estonia, Finland, France, Germany, Greece, Hungary, Italy, Latvia, Lithuania, Luxembourg, Netherlands, Norway, Poland, Portugal, Romania, Serbia, Slovakia, Slovenia, Spain, Sweden, Switzerland} European countries from the Ember Climate organization's website \footnote{https://ember-climate.org/data-catalogue/european-wholesale-electricity-price-data/}. The latter provides access to aggregated energy time series data, which was originally sourced from the European Network of Transmission System Operators for Electricity transparency platform \footnote{https://www.entsoe.eu/}. The wholesale electricity prices, which in our dataset are measured in euros per megawatt hour, are traded on the so-called spot markets. The latter enable the trade of the electricity output, i.e., the exchange of the energy between the electricity generators and suppliers. Therefore, the electricity price data used in our research is not related to the electricity prices paid by end-consumers, but rather by suppliers, who eventually sell the obtained electricity to their customers. Our dataset contains 5000 hours, i.e., approx. 208 days, which span from 2023-09-05 04:00 pm. to 2024-03-31 11:00 pm. according to UTC.\\

We argue that selecting a dataset (1) from the same data source and (2) across the European region offers several advantages for electricity price forecasting. First, it provides a unified view of the market across different closely integrated countries from a regulatory and reporting point of view, therefore avoiding potential distortions caused by the fragmentation often observed in other regions with different reporting and regulatory standards. Additionally, given that statistical methodologies and regulatory frameworks surrounding energy prices are largely harmonized at the European level, the dataset ensures a better comparability of data, thereby minimizing inconsistencies arising from national regulatory variations. Furthermore, the uniformity in data collection and reporting platform\footnote{https://transparency.entsoe.eu/} enhances the reliability and accuracy of data dynamics we observe. This makes it particularly suitable for evaluating market dynamics across Europe in a robust and standardized manner.

However, while the regulatory frameworks are mostly aligned, some exceptions exist, such as differences in bidding zones, price caps, and renewable energy integration across countries like Germany, Italy, and the Iberian Peninsula \cite{gonzalez2024mapping, Robinson2023}. These regulatory and market variations, rather than being drawbacks, offer a valuable opportunity to study how different forecasting models perform under varying conditions. By highlighting these differences, we can better understand how models respond to specific market complexities, such as fluctuations in renewable energy output or regulatory interventions. This enables a more precise comparison between forecasting model generations, revealing their strengths and weaknesses in adapting to both harmonized and unique market conditions across Europe.

\subsection*{Choosing the forecast time-steps}
We examine the correlational structure among the time steps using partial autocorrelation plots to identify suitable input sequence lengths for electricity price forecasting. Many studies on transformer-based frameworks for long-term time series forecasting commonly use an input sequence length of 96 time steps. \shortcite{Informer,Autoformer}.

\begin{figure}[H]
\centering
\includegraphics[scale=1.15]{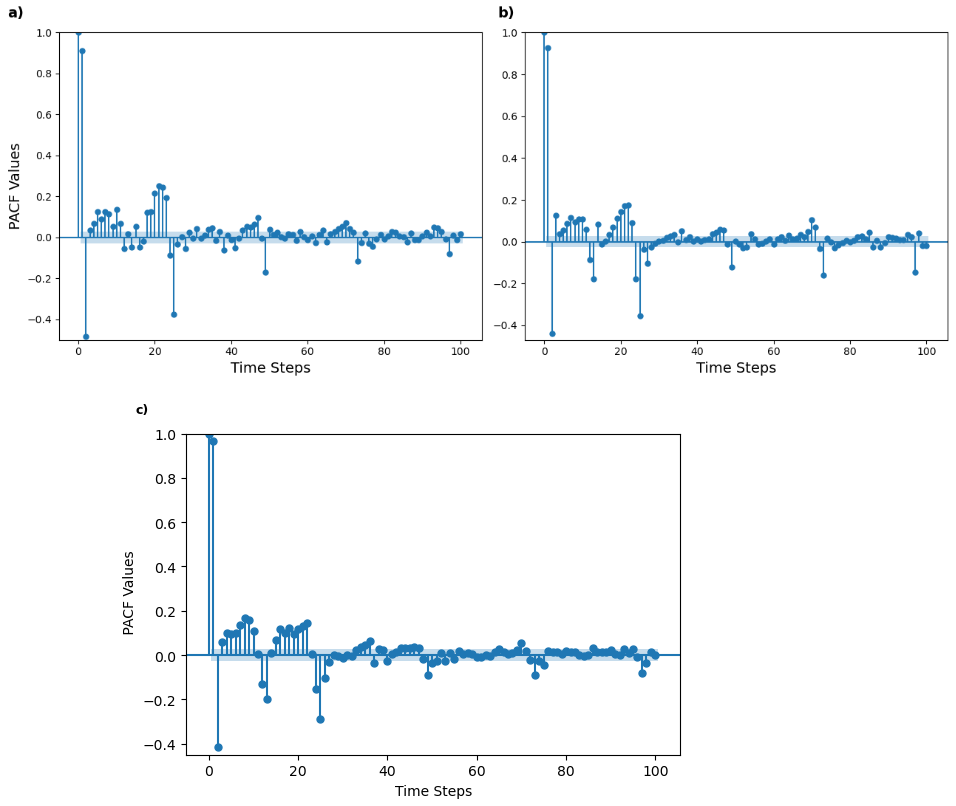}
\caption{PACF plots for a) Italy, b) Germany and c) Portugal.}
\label{fig:pacf}
\end{figure}

Figure \ref{fig:pacf} displays the Partial Autocorrelation Function (PACF) of three EU countries (Italy, Germany and Portugal) with 100 lags for pattern illustration purposes. The blue shaded area represents the critical region, where PACF coefficients are considered statistically insignificant. The PACF plots indicate that the coefficients with the highest absolute values are primarly concentrated within the first 24 lags. Although PACF values decrease with increasing lags, Figure \ref{fig:pacf} shows statistically significant correlations up to around 97–98 lags. Consequently, for our empirical analysis, we adopt the widely used input sequence length of 96 time steps.


\subsection{Experimental setup}
In this section, we detail the experimental design of our empirical research\footnote{The data and code can be found on Quantlet, under https://github.com/QuantLet/AI4EFin.git} and outline the evaluation criteria we use to measure the accuracy of the long-term predictions generated by the time series models included in our study.\\

Given our intra-day dataset, we define a step as one hour. Each model is trained to forecast the next 96 time steps using the past 96 steps of electricity prices. We use the default settings provided in the original codes retrieved from GitHub for each forecasting model. For example, the default number of training epochs for most models is set to 10. Given the high computational costs of training transformer-based models, we applied early stopping after a single epoch. The computational resources necessary to train the models were also the reason for taking the default hyper-parameters of the models instead of performing extensive hyper-parameter tuning. Thus, we did not use validation subsets, and early stopping was applied directly on the training data. Specifically, if we did not observe a minimum reduction of 0.01 units in the mean-squared error (MSE) loss function from one training epoch to the next, we terminated the training process. Additionally, in contrast to most studies in long-term time series forecasting, we used a walk-forward validation approach to assess the robustness of our forecasts on different test subsets, which has several advantages. First, unlike traditional train-test splits, walk-forward validation assesses model performance by incrementally moving the testing window forward in time, allowing the model to learn from past data and predict future, unseen data in a sequential manner. This approach, which involves retraining and testing the model on each new data point, reveals how well the model adapts to evolving patterns and trends. This, in turn, provides insights into the robustness and generalization capabilities of time series techniques. Additionally, walk-forward validation more closely simulates real-world forecasting scenarios by updating predictions as new data becomes available, ensuring that the model is tested under conditions that reflect practical deployment. \\

\begin{figure}[H]
\centering
\includegraphics[scale=0.45]{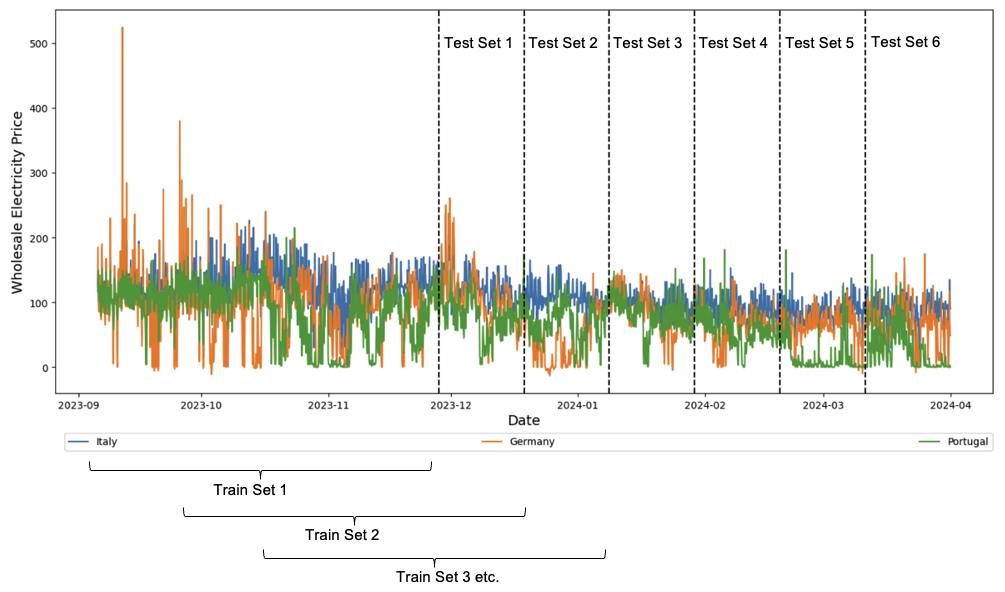}
\caption{Walk-forward validation with a rolling window based on the electricity prices of Italy, Germany and Portugal.}
\label{fig:trueTS}
\end{figure}

We split our electricity price time series into six train and test subsets and we exemplify the forward-walk validation split for Italy, Germany, and Portugal in Figure \ref{fig:trueTS}. Each train subset consists of a total of 2000 hours (83 days) and each test subset of 500 hours (20 days). The trainable parameters of neural-based techniques, except the pre-trained models, are transferred to every subsequent train subset and further tuned until convergence. The latter is reached when the early stopping criterion is fulfilled. While the pre-trained TimesFM and Chronos models, as well as the linear benchmark ARIMA, use univariate inputs (i.e., data from a single country), the other models are trained to "globally" forecast electricity prices across all 27 EU countries in our dataset. This refers to generating predictions for all input time series at once, without the need to train separate models in an iterative fashion in order to produce forecasts for each target time series. Therefore, the global prediction approach, which can be regarded as a variant of multi-target forecasting, is more scalable than the traditional iterative modelling approach. In addition to advantages related to efficiency, this technique uses multivariate input time series data. This implies that the inputs from all countries are considered simultaneously, so that electricity price data from one country can help improve the forecasts for another country (e.g., France and Germany), in case similar price development can be observed among the countries over time. We select SMAPE and RMSE as the evaluation criteria to measure the deviation of the predictions from the true values. Furthermore, the ranking of the models (Section \ref{subsec:results}), is based on the average of SMAPE and RMSE metrics. We standardize each train and test subset based on the mean and the standard deviation computed on the corresponding train subset. The evaluation of the predictions on the test subsets is performed on reverse re-scaled electricity prices.\\  




\subsection{Results}
\label{subsec:results}
In this section, we present the results from benchmarking the performance of 11 time series models using EU electricity price data. Firstly, we provide details about the best performing methods from each of the four generations of techniques. Secondly, In addition to ranking the models, we present insights about the performance of the models per country. Lastly, in the remaining part of the section we perform a robustness analysis of the results obtained from the walk-forward validation approach with a sliding window of 2000 training hours, i.e., 83 days.\\

Figure 4 depicts the average SMAPE (Symmetric Mean Absolute Percentage Error) across all time series models and reveals interesting clustering patterns among countries. This map highlights the presence of three distinct clusters based on the accuracy of the predictions.

\begin{figure}[h!] 
\centering
\includegraphics[scale=0.48]
{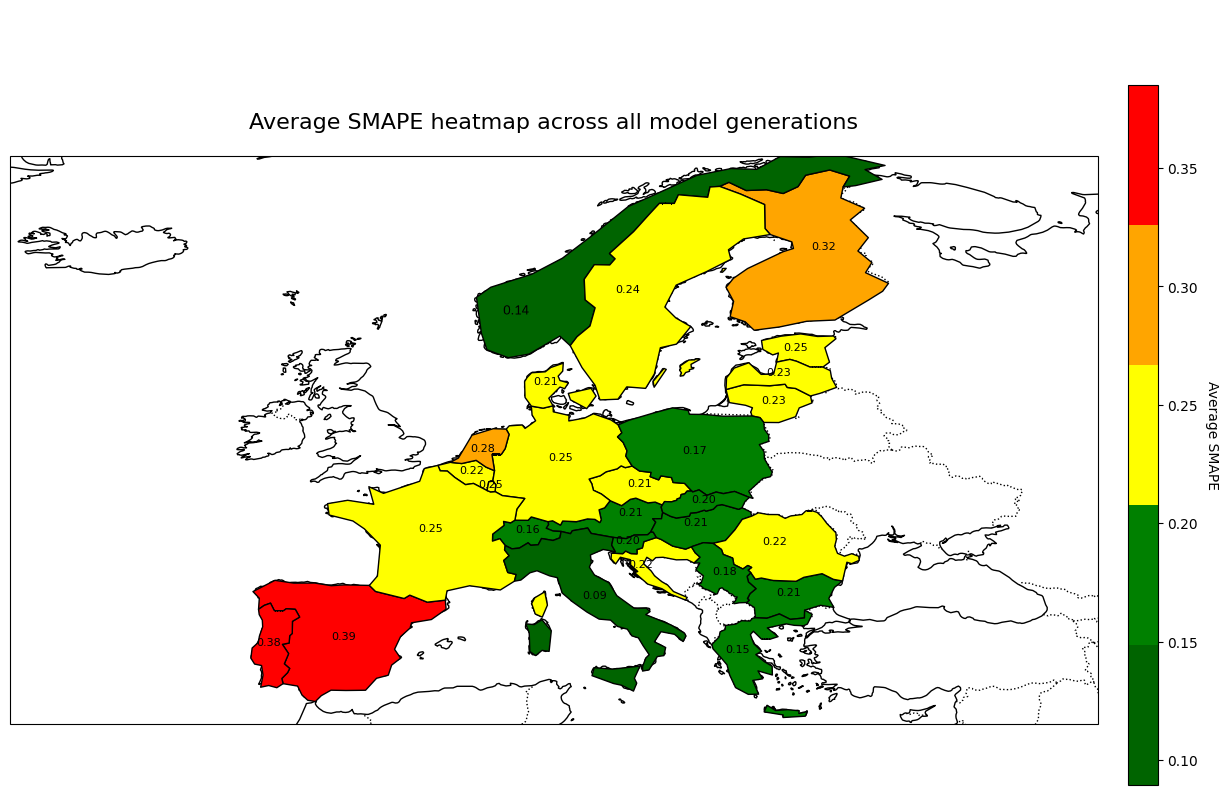}
\caption{Average SMAPE geomap achieved by all time series models for electricity price forecasting.}
\label{fig:geomap}
\end{figure}

The first cluster, characterized by the highest SMAPE values, includes Portugal, Spain, and Finland. This cluster indicates that time series models generally face greater challenges in forecasting electricity prices for these countries. The elevated errors observed can be attributed to several factors, such as more volatile electricity markets, specific market conditions given by unique regulatory environments, fluctuating demand patterns, or less robust historical data, all of which may complicate accurate price forecasting.

The second cluster contains countries such as Germany, France, and Sweden which display moderate SMAPE values (yellow). The moderate level errors suggest that electricity prices in these markets may exhibit more stable price patterns and better-quality data than those from the higher error cluster. On the other hand, they may still experience complex market dynamics, such as varying regulatory frameworks or diverse energy sources, which can introduce some level of prediction errors.

The third cluster, comprising countries such as Greece, Italy, Norway, Poland, and Switzerland, exhibits the lowest SMAPE values among all countries. The lower prediction errors in these countries can be attributed to several factors, including more stable and predictable electricity price patterns, high-quality historical data and less complex market conditions, allowing the models to deliver more accurate forecasts. The consistent trends and stable regulatory environments in these countries contribute to the models' effectiveness in capturing and predicting price movements accurately.

\begin{figure}
\begin{minipage}{0.5\textwidth}
\includegraphics[width=\textwidth]{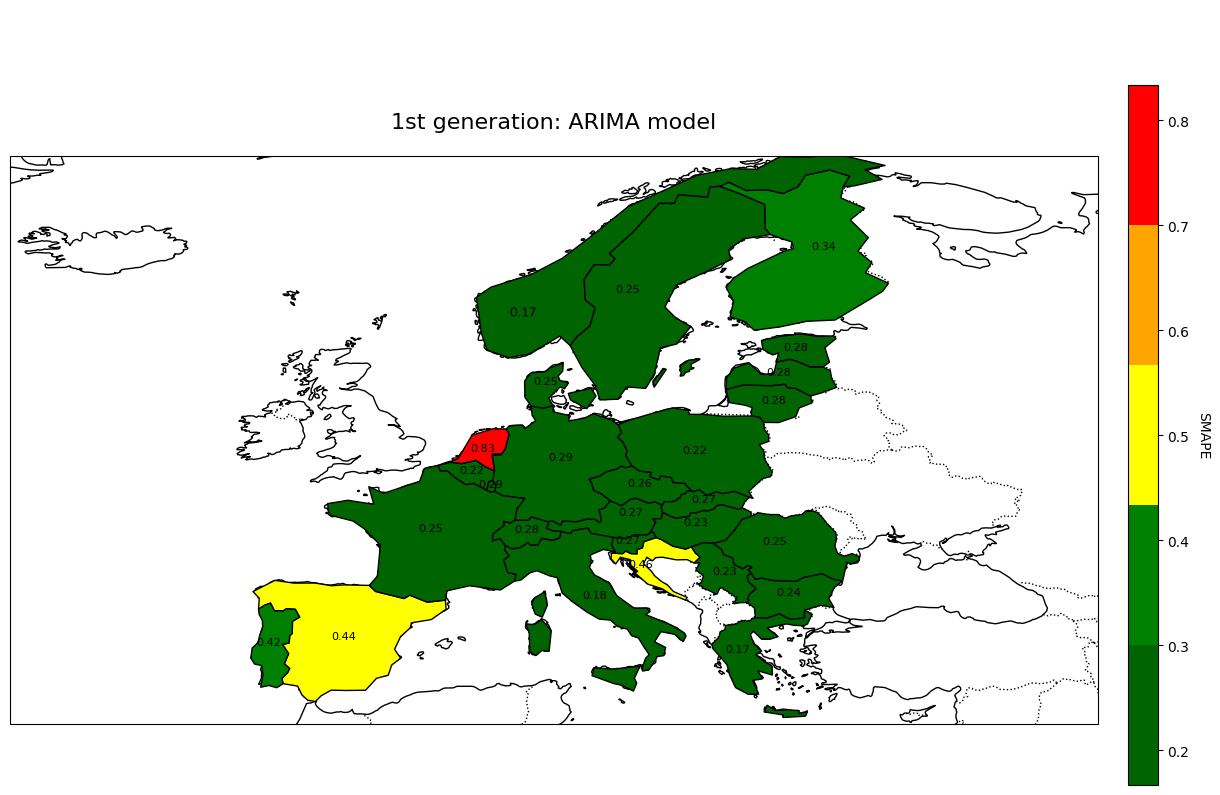}
\end{minipage}
\hfill
\begin{minipage}{0.5\textwidth}
\includegraphics[width=\textwidth]{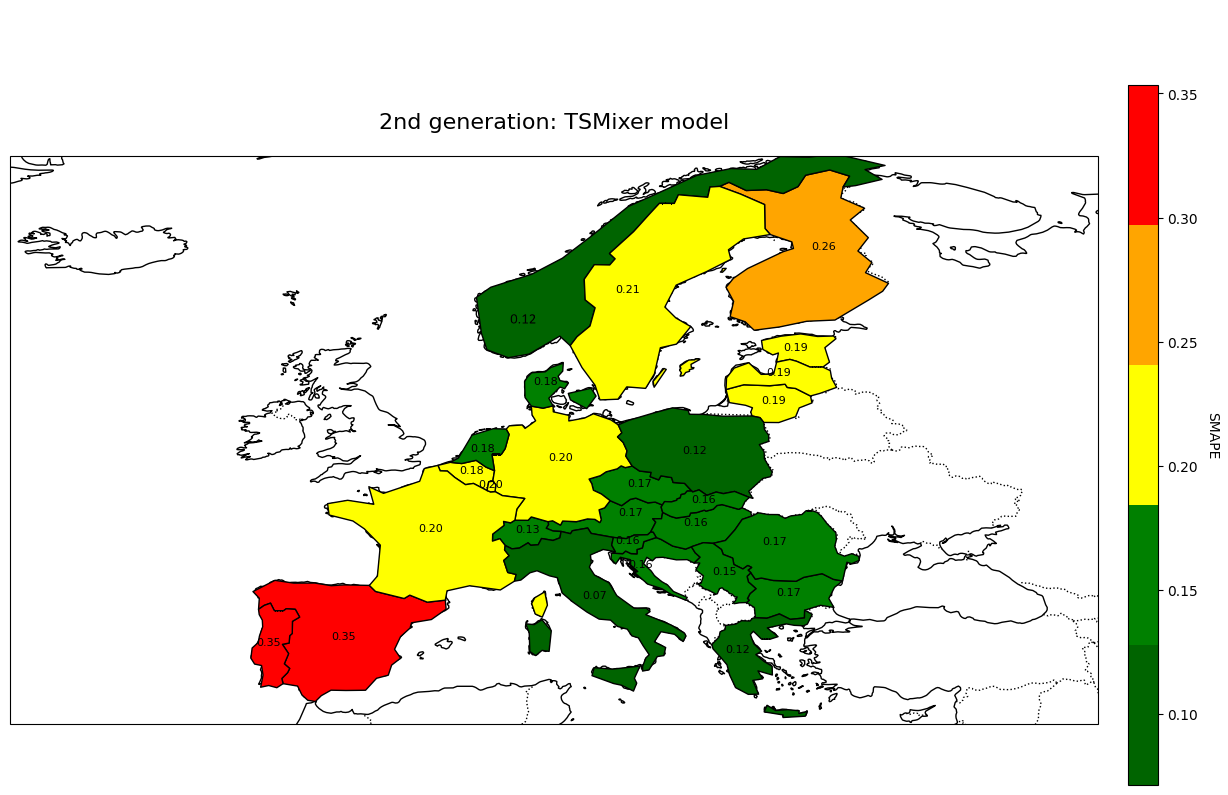}
\end{minipage}

\medskip 

\begin{minipage}{0.5\textwidth}
\includegraphics[width=\textwidth]{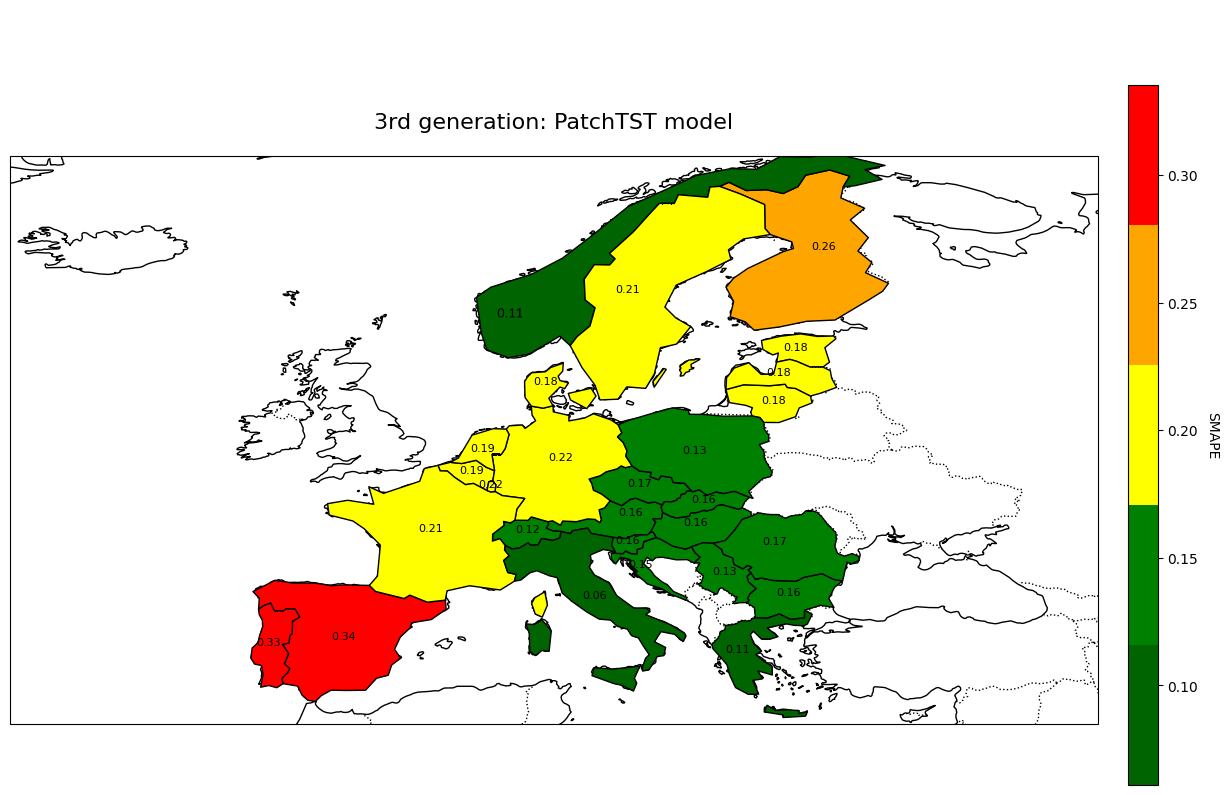}
\end{minipage}
\hfill
\begin{minipage}{0.5\textwidth}
\includegraphics[width=\textwidth]{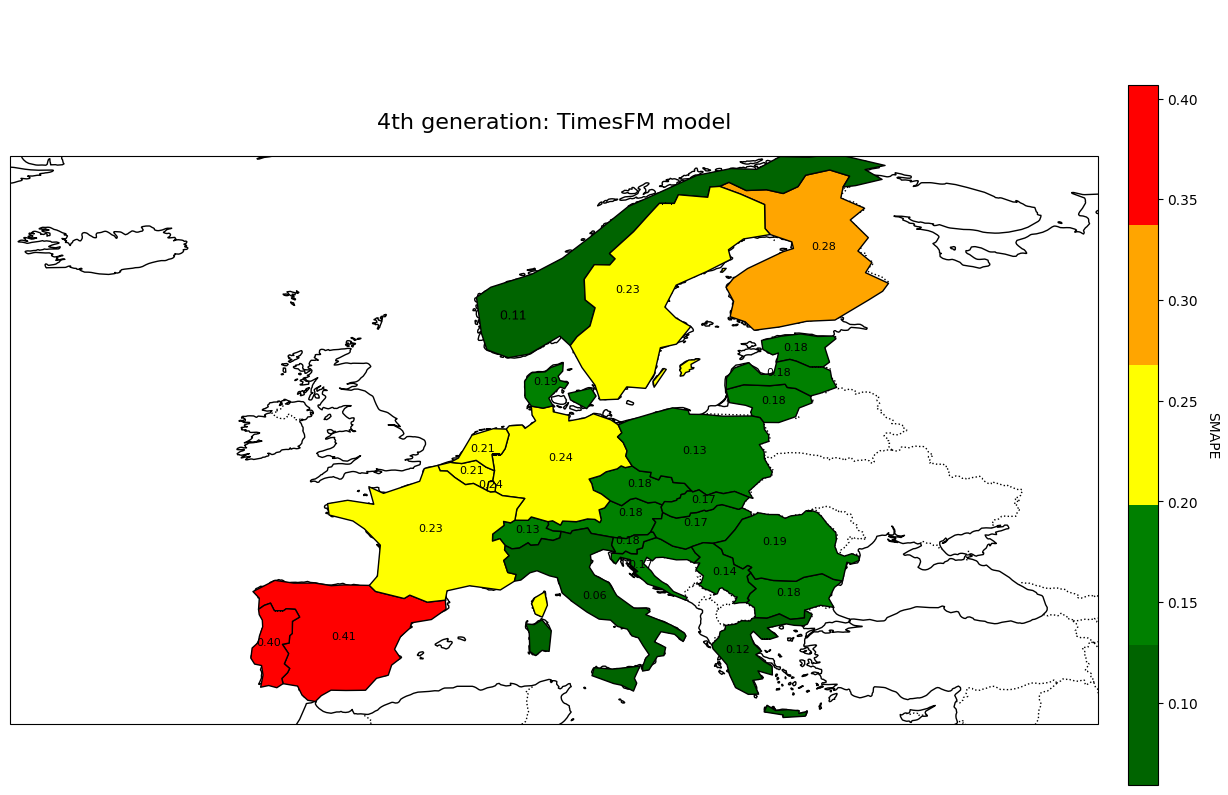}
\end{minipage}

\caption{Average SMAPE heatmaps with the best performing model per generation.}
\vspace{-10pt}
\label{fig:smape_models}
\end{figure}

One explanation for lower performance for Spain and Portugal is given by the "Iberian exception", which has allowed them to decouple electricity prices from natural gas prices, which has been a significant policy shift aimed at reducing consumer energy costs. This temporary mechanism was established in 2022 to counter the surge in energy prices due to external factors like the war in Ukraine. The cap on electricity prices, extended through 2024, is an intervention that reduces volatility in retail prices but can also introduce complexities—especially in a global forecasting framework— that models might struggle to predict accurately \cite{SerranoGonzalez2021, Abadie2021}. Furthermore, Spain and Portugal have relatively low interconnectivity with the broader European electricity grid and a high reliance on renewable energy sources. While beneficial for sustainability, this situation can lead to more erratic price fluctuations, as renewable energy introduces variability based on supply conditions. These factors, combined with differing regulatory structures from other European countries, make it more difficult for models to capture and predict electricity price trends accurately \cite{Robinson2023}. Thus, the combination of regulatory interventions like price caps and market-specific factors likely explains why forecasting models have encountered more difficulty in these regions.

In Figure 5, we plot the heatmaps for the top performing models in each of the four generations. We note similar results for the data from the clusters of countries identified in Figure 4. However, one exception is given by the ARIMA model, which exhibits reduced performance in time series forecasting. This discrepancy highlights the limitations of linear econometric models in adapting to the unique challenges posed by contemporary electricity markets. In contrast, the other more advanced models from subsequent generations, show similar dynamics thereby reflecting improved adaptability and accuracy in capturing the intricacies of price fluctuations across the different clusters of countries.


In addition to analyzing the results in terms of SMAPE, we rank the time series models included in our empirical research based on the performance indicator:

\begin{figure}[H] 
\centering
\begin{subfigure}{\textwidth} 
    \centering
    \includegraphics[scale=0.3]{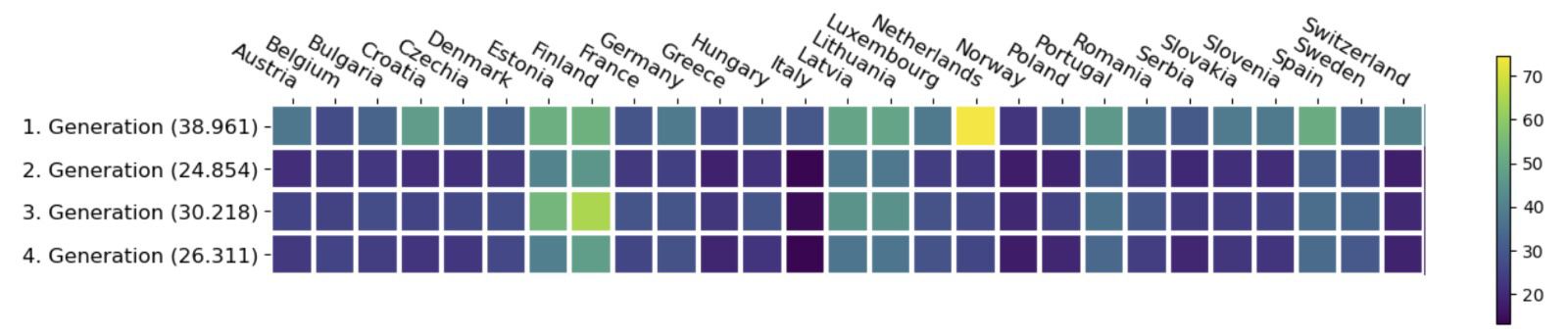} 
    \caption{Heatmap of average performance indicator values achieved by model generation.}
    \label{fig:topfigure}
\end{subfigure}

\vspace{0.5cm} 

\begin{subfigure}{\textwidth} 
    \centering
    \includegraphics[scale=0.65]{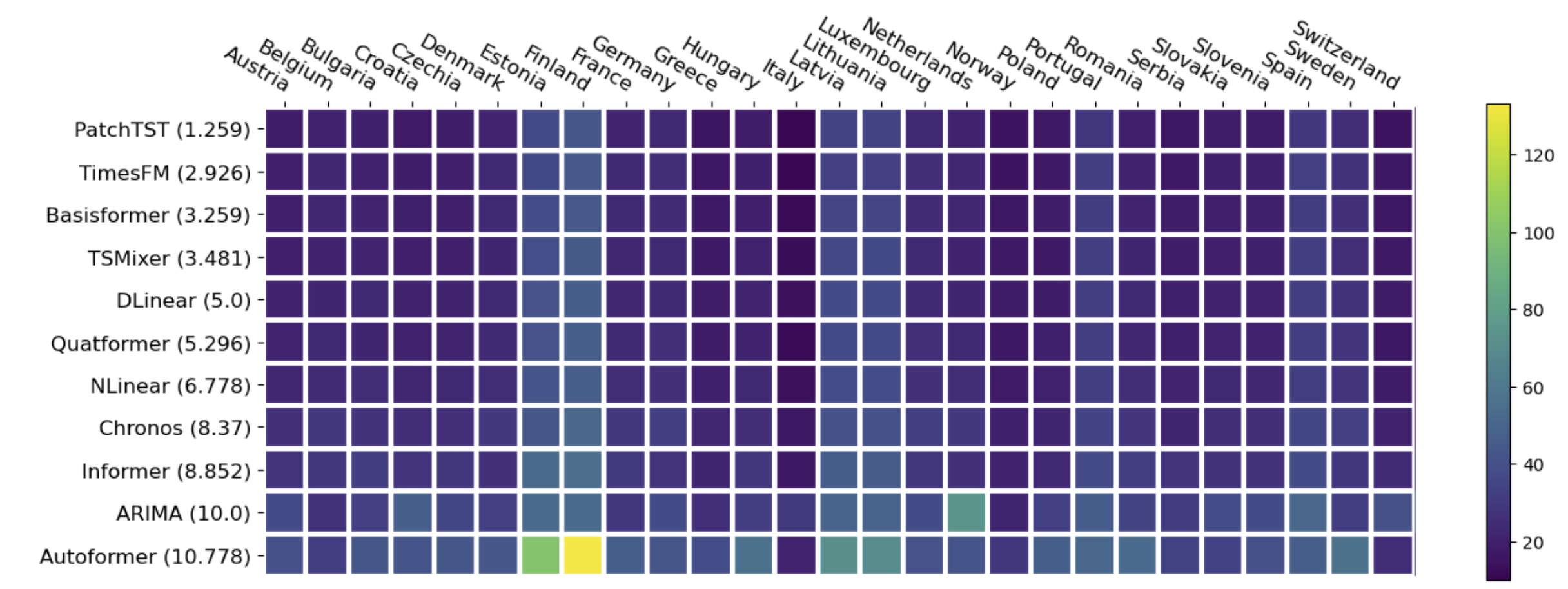} 
    \caption{Heatmap of average performance indicator values achieved by each time series model per EU country.}
    \label{fig:heatmapSMAPERMSE}
\end{subfigure}

\caption{Heatmap of avearge performance indicators per model generations (a) and for each model (b)}
\label{fig:combined_figure}
\end{figure}

\begin{equation}
    \hfill
    \frac{100 \cdot SMAPE \plus RMSE}{2}
    \hfill
\end{equation}

While SMAPE evaluates the percentual deviation of the forecasts from the true values, RMSE measures the error in the original scale of the target variables. Thus, by incorporating both metrics in our average performance indicator, which we use in order to rank the models, we aim at assessing the performance in a robust way from two different perspectives. The models on the y-axis of the heatmap in Figure \ref{fig:heatmapSMAPERMSE} are sorted in an ascending order of the average rank, produced by the time series techniques on each of the 27 EU countries. In Appendix \ref{sec:Appendix1}, we provide more details about the ranking of each model per country. The first two best performing models in Figure \ref{fig:heatmapSMAPERMSE} share certain algorithmic similarities. Both models are transformer-based, and extract latent features from univariate time series. While PatchtTST makes use of channel-independence to reduce the number of trainable paremeters, the weights are transfered to each channel in the multivariate time series inputs during the training process. TimesFM, which represents a foundation model, can generate forecasts only from univariate inputs. However, in contrast to PatchTST, TimesFM generates forecasts without being trained on a specific dataset of interest, as the model was pre-trained on a large corpus of univarite time series datasets. The fact that TimesFM outperforms 9 out 11 models on the challenging task of long-term forecasting of electricity prices highlights the potential for a significant reduction in the computational cost associated with training of traditional forecasting techniques.\\

Although PatchTST and TimesFM outperm the other models included in our empirical research, the heatmap in Figure \ref{fig:heatmapSMAPERMSE} shows that more or less all forecasting techniques produce less accurate forecasts of the next 96 hours on, e.g., Finland, Spain, and Portugal, than on most of the remaining countries in our dataset. Finland, Spain, and Portugal represent countries, which are characterized by more volatile and unpredictable electricity market conditions, driven by several factors. Firstly, these countries have lower market integration with the broader European grid, which limits the ability to balance supply and demand fluctuation through cross-border electricity flows. Secondly, they have a higher reliance on renewable energy sources that introduce variability into price patterns \cite{bento2024soaring}. Additionally, regulatory differences, further complicate the models' ability to capture these dynamics accurately, as previously mentioned. 
Furthermore, Finland's electricity market has become one of the most volatile in Europe due to various factors like unpredictable weather, renewable energy fluctuations, and outages in nuclear facilities \footnote{Source: Energiateollisuus \url{https://energia.fi/wp-content/uploads/2024/01/Electricity-price-statistics-2023.pdf}}. The combination of nuclear power outages and variable renewable energy sources, such as wind, has led to significant price swings, including periods of negative prices. \footnote{Source: Montel Energy \url{https://montelgroup.com/updates-and-insights/finland-europes-most-volatile-short-term-electricity-market?trk=public_post_comment-text}}.\\

\begin{figure}[H]
\centering
\includegraphics[scale=0.5]{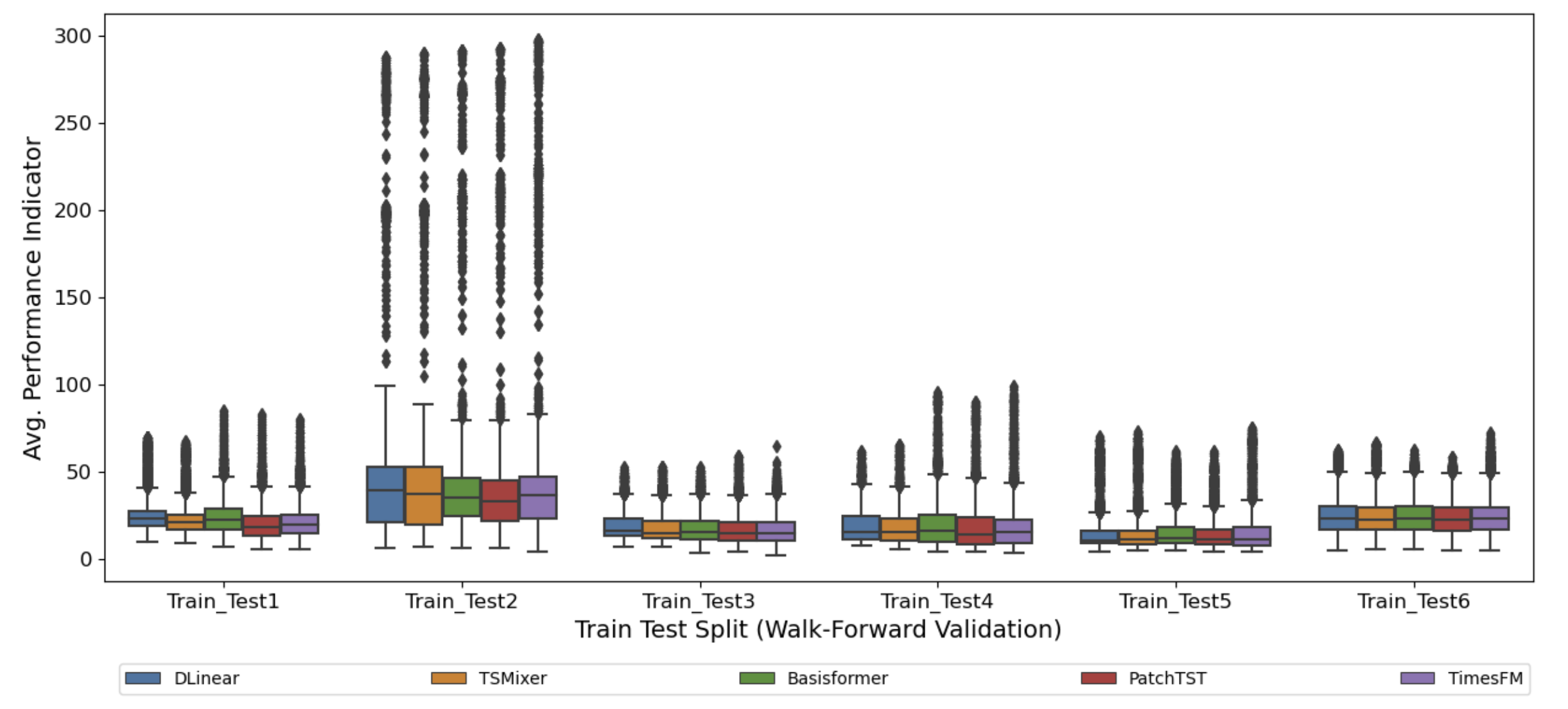}
\caption{Robustness analysis of top five best performing time series models using walk-forward validation.}
\label{fig:boxplot}
\end{figure}

We next turn our attention model performance for the first five spots, i.e., PatchTST, TimesFM, TSMixer, Basisformer and DLinear. The conditional boxplots in Figure \ref{fig:boxplot} plot the model performance on each of the six test subsets.  

The top five performing models demonstrate generally comparable results across the test subsets, with the notable exception of the second test subset. In this subset, the interquartile range (IQR) of the boxplots reveals a greater spread of the prediction errors, and the presence of extreme outliers indicates that the models are facing difficulties with predictions specific to this subset. To provide further insights, we focus on the results from Germany by analyzing the model predictions across the first, second, and final test subsets.


\begin{figure}[H]
\centering
\includegraphics[scale=0.47]{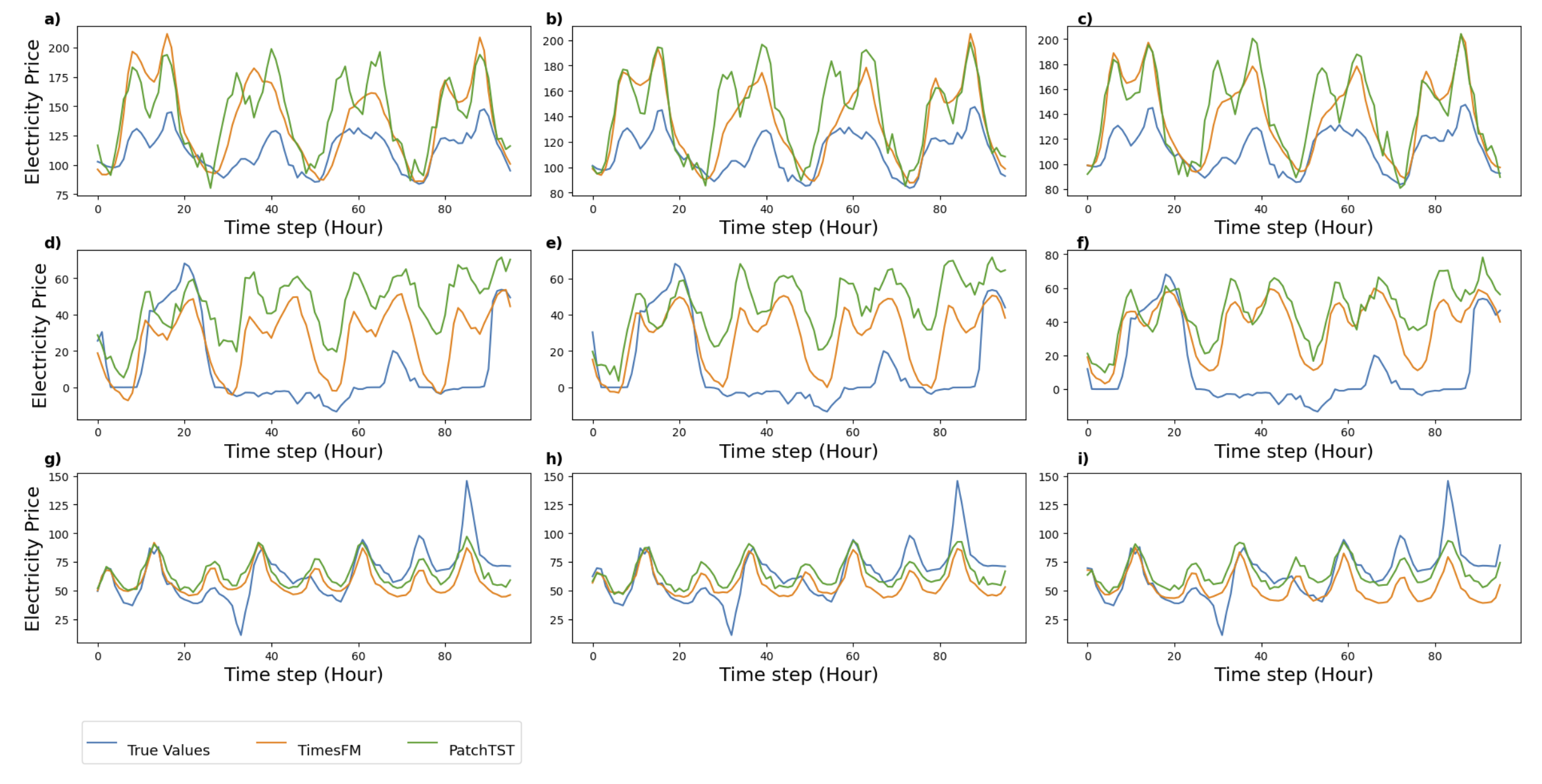}
\caption{Actual vs. Predicted values obtained from PatchTST and TimesFM on Germany, a), b), c): test subset 1 with the first three sequence samples, d), e), f): test subset 2 with the first three sequence samples, g), h), i): test subset 6 with the first three sequence samples,}
\label{fig:tscomparison}
\end{figure}

The time series plots show that both TimesFM and PatchTST struggle to capture the pattern of very low positive as well as negative electricity prices, particularly on the second subset of the test data in Figures \ref{fig:tscomparison}d), \ref{fig:tscomparison}e) and \ref{fig:tscomparison}f). We attribute the under-performance to higher price volatility during that snippet of data. However, the performance significantly improves within the last subset, where the absence of sudden price fluctuations allows both models to capture the overall dynamics more effectively. When the true electricity prices follow a repetitive or seasonal pattern, TimesFM and PatchTST are able to predict future price movements over the next 96 time steps with higher accuracy. This suggests that the models are better suited for handling stable, cyclical trends than they are at responding to sharp, unpredictable price shifts.


\section{Conclusion}
In this study, we conducted a benchmarking analysis of 11 time series models from four generations of modeling techniques, focusing on forecasting a 96-hour ahead time period of EU electricity prices. Our empirical findings indicate that forecasting models that use transformer-based architectures and Large Language Models (LLMs) have superior performance in capturing the complexities inherent in high-frequency electricity time series. The attention mechanisms that these models encompass make them suitable to model long-range dependencies in time series data, thus improving their explanatory power as compared to other traditional methods. 

Our analysis indicates that transformer-based PatchTST and Basisformer are among the most effective techniques for forecasting electricity prices, with their integration of trend and seasonal adjustments contributing to enhanced predictive accuracy. LLM-based models such as TimesFM offer a distinct advantage due to their extensive pre-training on diverse datasets, including past text corpora. This enables them to generalize effectively and provide accurate forecasts without the need to train LLMs on task-specific data. In addition to advantages related to reduction in the computational cost required to train models from scratch, the zero-shot forecasting capabilities of LLMs makes them particularly beneficial in settings with limited data availability. 

Our study highlights the importance of benchmarking various families of time series techniques for forecasting datasets that have not been previously examined in empirical analyses. Specifically, our results indicate that two of the top three models in our benchmarks are transformer-based, while the third is a foundation model (FM), underscoring the importance of incorporating advanced AI tools like transformers and foundation models into energy price forecasting. Including different modeling techniques allows for a more comprehensive assessment of forecasting accuracy and facilitates comparisons across approaches, potentially providing insights into how various models capture the inherent complexities and non-linear patterns typical of energy time-series data. 
This study contributes to the field of energy price forecasting by benchmarking the performance of different time series models, particularly transformer-based architectures and Large Language Models, against deep learning and linear econometric techniques. The results show their superior accuracy and flexibility in handling the complexities of high-frequency electricity price data.\\

Future research could explore the performance of the four model families on synthetic data generated with varying levels of complexity, such as multiple seasonal components, alternating upward and downward trends with reversals and plateau intervals of different lengths. This approach would help identify challenging scenarios that could expose performance limitations. Additionally, exploring hybrid approaches, which combine the strengths of different architectures could improve the expressive power of long-term forecasting techniques for time-series data. Given the algorithmic complexity of models of third- and fourth-generation models, it would be useful to investigate which specific components contribute most to long-term prediction accuracy, thereby increasing the transparency and interpretability of these models. This in turn could facilitate the adoption of these techniques in practical decision-making processes. Furthermore, future research could leverage LLMs and other AI models for risk measurement, employing metrics such as Value at Risk. Integrating these models into risk management frameworks would enable more accurate and comprehensive assessments of risks associated with energy price fluctuations, further supporting decision-making in the energy sector.

\section*{Acknowledgments}
This paper is supported through the European Cooperation in Science \& Technology COST Action grant CA19130 - Fintech and Artificial Intelligence in Finance - Towards a transparent financial industry; the project ``IDA Institute of Digital Assets'', CF166/15.11.2022, contract number CN760046/ 23.05.2023; the project ``AI for Energy Finance (AI4EFin)'', CF162/15.11.2022, contract number CN760048/23.05.2023, financed under the Romania’s National Recovery and Resilience Plan, Apel nr. PNRR-III-C9-2022-I8; and the Marie Skłodowska-Curie Actions under the European Union's Horizon Europe research and innovation program for the Industrial Doctoral Network on Digital Finance, acronym: DIGITAL, Project No. 101119635.

\newpage
\bibliographystyle{apacite}

\begin{thebibliography}{}

\bibitem [\protect \citeauthoryear {%
Abadie%
}{%
Abadie%
}{%
{\protect \APACyear {2021}}%
}]{%
Abadie2021}
\APACinsertmetastar {%
Abadie2021}%
\begin{APACrefauthors}%
Abadie, L\BPBI M.%
\end{APACrefauthors}%
\unskip\
\newblock
\APACrefYearMonthDay{2021}{}{}.
\newblock
{\BBOQ}\APACrefatitle {{Energy} {Market} {Prices} in {Times} of {COVID-19}: {The} {Case} of {Electricity} and {Natural Gas} in {Spain}} {{Energy} {Market} {Prices} in {Times} of {COVID-19}: {The} {Case} of {Electricity} and {Natural Gas} in {Spain}}.{\BBCQ}
\newblock
\APACjournalVolNumPages{Energies}{14}{6}{1632}.
\newblock
\APACrefnote{Submission received: 10 February 2021 / Revised: 9 March 2021 / Accepted: 11 March 2021 / Published: 15 March 2021}
\newblock
\begin{APACrefDOI} \doi{10.3390/en14061632} \end{APACrefDOI}
\PrintBackRefs{\CurrentBib}

\bibitem [\protect \citeauthoryear {%
Ansari%
\ \protect \BOthers {.}}{%
Ansari%
\ \protect \BOthers {.}}{%
{\protect \APACyear {2024}}%
}]{%
ansari_chronos_2024}
\APACinsertmetastar {%
ansari_chronos_2024}%
\begin{APACrefauthors}%
Ansari, A\BPBI F.%
, Stella, L.%
, Turkmen, C.%
, Zhang, X.%
, Mercado, P.%
, Shen, H.%
\BDBL {}Wang, Y.%
\end{APACrefauthors}%
\unskip\
\newblock
\APACrefYearMonthDay{2024}{{\APACmonth{05}}}{}.
\newblock
\APACrefbtitle {Chronos: {Learning} the {Language} of {Time} {Series}.} {Chronos: {Learning} the {Language} of {Time} {Series}.}
\newblock
\APACaddressPublisher{}{arXiv}.
\newblock
\begin{APACrefURL} [{2024-06-26}]\url{http://arxiv.org/abs/2403.07815} \end{APACrefURL}
\newblock
\APACrefnote{arXiv:2403.07815 [cs]}
\newblock
\begin{APACrefDOI} \doi{10.48550/arXiv.2403.07815} \end{APACrefDOI}
\PrintBackRefs{\CurrentBib}

\bibitem [\protect \citeauthoryear {%
Bento%
, Mariano%
, Carvalho%
, Calado%
\BCBL {}\ \BBA {} Pombo%
}{%
Bento%
\ \protect \BOthers {.}}{%
{\protect \APACyear {2024}}%
}]{%
bento2024soaring}
\APACinsertmetastar {%
bento2024soaring}%
\begin{APACrefauthors}%
Bento, P.%
, Mariano, S.%
, Carvalho, P.%
, Calado, M.%
\BCBL {}\ \BBA {} Pombo, J.%
\end{APACrefauthors}%
\unskip\
\newblock
\APACrefYearMonthDay{2024}{}{}.
\newblock
{\BBOQ}\APACrefatitle {Soaring electricity prices in the day-ahead {Iberian} market: policy insights, regulatory challenges and lack of system flexibility} {Soaring electricity prices in the day-ahead {Iberian} market: policy insights, regulatory challenges and lack of system flexibility}.{\BBCQ}
\newblock
\APACjournalVolNumPages{International Journal of Energy Sector Management}{18}{2}{312--333}.
\newblock
\begin{APACrefURL} \url{https://doi.org/10.1108/IJESM-07-2022-0007} \end{APACrefURL}
\newblock
\begin{APACrefDOI} \doi{10.1108/IJESM-07-2022-0007} \end{APACrefDOI}
\PrintBackRefs{\CurrentBib}

\bibitem [\protect \citeauthoryear {%
Bierbrauer%
, Menn%
, Rachev%
\BCBL {}\ \BBA {} Trück%
}{%
Bierbrauer%
\ \protect \BOthers {.}}{%
{\protect \APACyear {2007}}%
}]{%
bierbrauer2007spot}
\APACinsertmetastar {%
bierbrauer2007spot}%
\begin{APACrefauthors}%
Bierbrauer, M.%
, Menn, C.%
, Rachev, S\BPBI T.%
\BCBL {}\ \BBA {} Trück, S.%
\end{APACrefauthors}%
\unskip\
\newblock
\APACrefYearMonthDay{2007}{}{}.
\newblock
{\BBOQ}\APACrefatitle {Spot and derivative pricing in the {EEX} power market} {Spot and derivative pricing in the {EEX} power market}.{\BBCQ}
\newblock
\APACjournalVolNumPages{Journal of Banking \& Finance}{31}{}{3462--3485}.
\PrintBackRefs{\CurrentBib}

\bibitem [\protect \citeauthoryear {%
Cantillo-Luna%
, Moreno-Chuquen%
, Lopez-Sotelo%
\BCBL {}\ \BBA {} Celeita%
}{%
Cantillo-Luna%
\ \protect \BOthers {.}}{%
{\protect \APACyear {2023}}%
}]{%
37_Cantillo-Luna}
\APACinsertmetastar {%
37_Cantillo-Luna}%
\begin{APACrefauthors}%
Cantillo-Luna, S.%
, Moreno-Chuquen, R.%
, Lopez-Sotelo, J.%
\BCBL {}\ \BBA {} Celeita, D.%
\end{APACrefauthors}%
\unskip\
\newblock
\APACrefYearMonthDay{2023}{}{}.
\newblock
{\BBOQ}\APACrefatitle {An {Intra}-{Day} {Electricity} {Price} {Forecasting} {Based} on a {Probabilistic} {Transformer} {Neural} {Network} {Architecture}} {An {Intra}-{Day} {Electricity} {Price} {Forecasting} {Based} on a {Probabilistic} {Transformer} {Neural} {Network} {Architecture}}.{\BBCQ}
\newblock
\APACjournalVolNumPages{Energies}{16}{19}{}.
\newblock
\begin{APACrefURL} \url{https://www.mdpi.com/1996-1073/16/19/6767} \end{APACrefURL}
\newblock
\begin{APACrefDOI} \doi{10.3390/en16196767} \end{APACrefDOI}
\PrintBackRefs{\CurrentBib}

\bibitem [\protect \citeauthoryear {%
S\BHBI A.~Chen%
, Li%
, Yoder%
, Arik%
\BCBL {}\ \BBA {} Pfister%
}{%
S\BHBI A.~Chen%
\ \protect \BOthers {.}}{%
{\protect \APACyear {2023}}%
}]{%
TSMixer}
\APACinsertmetastar {%
TSMixer}%
\begin{APACrefauthors}%
Chen, S\BHBI A.%
, Li, C\BHBI L.%
, Yoder, N.%
, Arik, S\BPBI O.%
\BCBL {}\ \BBA {} Pfister, T.%
\end{APACrefauthors}%
\unskip\
\newblock
\APACrefYearMonthDay{2023}{}{}.
\newblock
\APACrefbtitle {{TSMixer}: {An} {All}-{MLP} {Architecture} for {Time} {Series} {Forecasting}.} {{TSMixer}: {An} {All}-{MLP} {Architecture} for {Time} {Series} {Forecasting}.}
\newblock
\begin{APACrefURL} \url{https://arxiv.org/abs/2303.06053} \end{APACrefURL}
\PrintBackRefs{\CurrentBib}

\bibitem [\protect \citeauthoryear {%
W.~Chen%
\ \protect \BOthers {.}}{%
W.~Chen%
\ \protect \BOthers {.}}{%
{\protect \APACyear {2022}}%
}]{%
Quatformer}
\APACinsertmetastar {%
Quatformer}%
\begin{APACrefauthors}%
Chen, W.%
, Wang, W.%
, Peng, B.%
, Wen, Q.%
, Zhou, T.%
\BCBL {}\ \BBA {} Sun, L.%
\end{APACrefauthors}%
\unskip\
\newblock
\APACrefYearMonthDay{2022}{}{}.
\newblock
{\BBOQ}\APACrefatitle {{Learning to rotate: Quaternion transformer for complicated periodical time series forecasting}} {{Learning to rotate: Quaternion transformer for complicated periodical time series forecasting}}.{\BBCQ}
\newblock
\BIn{} \APACrefbtitle {{Proceedings of Conference on Knowledge Discovery and Data Mining}.} {{Proceedings of Conference on Knowledge Discovery and Data Mining}.}
\PrintBackRefs{\CurrentBib}

\bibitem [\protect \citeauthoryear {%
Das%
, Kong%
, Sen%
\BCBL {}\ \BBA {} Zhou%
}{%
Das%
\ \protect \BOthers {.}}{%
{\protect \APACyear {2024}}%
}]{%
das_decoder-only_2024}
\APACinsertmetastar {%
das_decoder-only_2024}%
\begin{APACrefauthors}%
Das, A.%
, Kong, W.%
, Sen, R.%
\BCBL {}\ \BBA {} Zhou, Y.%
\end{APACrefauthors}%
\unskip\
\newblock
\APACrefYearMonthDay{2024}{{\APACmonth{04}}}{}.
\newblock
\APACrefbtitle {A decoder-only foundation model for time-series forecasting.} {A decoder-only foundation model for time-series forecasting.}
\newblock
\APACaddressPublisher{}{arXiv}.
\newblock
\begin{APACrefURL} [{2024-06-26}]\url{http://arxiv.org/abs/2310.10688} \end{APACrefURL}
\newblock
\APACrefnote{arXiv:2310.10688 [cs]}
\newblock
\begin{APACrefDOI} \doi{10.48550/arXiv.2310.10688} \end{APACrefDOI}
\PrintBackRefs{\CurrentBib}

\bibitem [\protect \citeauthoryear {%
Gabrielli%
, Wüthrich%
, Blume%
\BCBL {}\ \BBA {} Sansavini%
}{%
Gabrielli%
\ \protect \BOthers {.}}{%
{\protect \APACyear {2022}}%
}]{%
gabrielli2022data}
\APACinsertmetastar {%
gabrielli2022data}%
\begin{APACrefauthors}%
Gabrielli, P.%
, Wüthrich, M.%
, Blume, S.%
\BCBL {}\ \BBA {} Sansavini, G.%
\end{APACrefauthors}%
\unskip\
\newblock
\APACrefYearMonthDay{2022}{}{}.
\newblock
{\BBOQ}\APACrefatitle {Data-driven modeling for long-term electricity price forecasting} {Data-driven modeling for long-term electricity price forecasting}.{\BBCQ}
\newblock
\APACjournalVolNumPages{Energy}{}{}{}.
\newblock
\begin{APACrefURL} \url{https://doi.org/10.1016/j.energy.2022.123107} \end{APACrefURL}
\newblock
\begin{APACrefDOI} \doi{10.1016/j.energy.2022.123107} \end{APACrefDOI}
\PrintBackRefs{\CurrentBib}

\bibitem [\protect \citeauthoryear {%
Ghoddusi%
, Creamer%
\BCBL {}\ \BBA {} Rafizadeh%
}{%
Ghoddusi%
\ \protect \BOthers {.}}{%
{\protect \APACyear {2019}}%
}]{%
GHODDUSI2019}
\APACinsertmetastar {%
GHODDUSI2019}%
\begin{APACrefauthors}%
Ghoddusi, H.%
, Creamer, G\BPBI G.%
\BCBL {}\ \BBA {} Rafizadeh, N.%
\end{APACrefauthors}%
\unskip\
\newblock
\APACrefYearMonthDay{2019}{}{}.
\newblock
{\BBOQ}\APACrefatitle {Machine learning in energy economics and finance: A review} {Machine learning in energy economics and finance: A review}.{\BBCQ}
\newblock
\APACjournalVolNumPages{Energy Economics}{81}{}{709-727}.
\newblock
\begin{APACrefURL} \url{https://www.sciencedirect.com/science/article/pii/S0140988319301513} \end{APACrefURL}
\newblock
\begin{APACrefDOI} \doi{https://doi.org/10.1016/j.eneco.2019.05.006} \end{APACrefDOI}
\PrintBackRefs{\CurrentBib}

\bibitem [\protect \citeauthoryear {%
González%
\ \BBA {} Álvarez Alonso%
}{%
González%
\ \BBA {} Álvarez Alonso%
}{%
{\protect \APACyear {2021}}%
}]{%
SerranoGonzalez2021}
\APACinsertmetastar {%
SerranoGonzalez2021}%
\begin{APACrefauthors}%
González, J\BPBI S.%
\BCBT {}\ \BBA {} Álvarez Alonso, C.%
\end{APACrefauthors}%
\unskip\
\newblock
\APACrefYearMonthDay{2021}{}{}.
\newblock
{\BBOQ}\APACrefatitle {{Industrial electricity prices in Spain: A discussion in the context of the European internal energy market}} {{Industrial electricity prices in Spain: A discussion in the context of the European internal energy market}}.{\BBCQ}
\newblock
\APACjournalVolNumPages{Energy Policy}{148}{}{111930}.
\newblock
\begin{APACrefDOI} \doi{10.1016/j.enpol.2020.111930} \end{APACrefDOI}
\PrintBackRefs{\CurrentBib}

\bibitem [\protect \citeauthoryear {%
González-de Miguel%
, van Wunnik%
\BCBL {}\ \BBA {} Sumper%
}{%
González-de Miguel%
\ \protect \BOthers {.}}{%
{\protect \APACyear {2024}}%
}]{%
gonzalez2024mapping}
\APACinsertmetastar {%
gonzalez2024mapping}%
\begin{APACrefauthors}%
González-de Miguel, C.%
, van Wunnik, L.%
\BCBL {}\ \BBA {} Sumper, A.%
\end{APACrefauthors}%
\unskip\
\newblock
\APACrefYearMonthDay{2024}{}{}.
\newblock
{\BBOQ}\APACrefatitle {{Mapping the Wholesale Day-Ahead Market Effects of the Gas Subsidy in the Iberian Exception}} {{Mapping the Wholesale Day-Ahead Market Effects of the Gas Subsidy in the Iberian Exception}}.{\BBCQ}
\newblock
\APACjournalVolNumPages{Energies}{17}{13}{3102}.
\newblock
\begin{APACrefURL} \url{https://doi.org/10.3390/en17133102} \end{APACrefURL}
\newblock
\begin{APACrefDOI} \doi{10.3390/en17133102} \end{APACrefDOI}
\PrintBackRefs{\CurrentBib}

\bibitem [\protect \citeauthoryear {%
Gruver%
, Finzi%
, Qiu%
\BCBL {}\ \BBA {} Wilson%
}{%
Gruver%
\ \protect \BOthers {.}}{%
{\protect \APACyear {{\protect \bibnodate {}}}}%
}]{%
gruver_large_nodate}
\APACinsertmetastar {%
gruver_large_nodate}%
\begin{APACrefauthors}%
Gruver, N.%
, Finzi, M.%
, Qiu, S.%
\BCBL {}\ \BBA {} Wilson, A\BPBI G.%
\end{APACrefauthors}%
\unskip\
\newblock
\APACrefYearMonthDay{{\protect \bibnodate {}}}{}{}.
\newblock
{\BBOQ}\APACrefatitle {Large {Language} {Models} {Are} {Zero}-{Shot} {Time} {Series} {Forecasters}} {Large {Language} {Models} {Are} {Zero}-{Shot} {Time} {Series} {Forecasters}}.{\BBCQ}
\newblock

\PrintBackRefs{\CurrentBib}

\bibitem [\protect \citeauthoryear {%
Gueta%
, Feder%
, Gekhman%
, Goldstein%
\BCBL {}\ \BBA {} Reichart%
}{%
Gueta%
\ \protect \BOthers {.}}{%
{\protect \APACyear {2024}}%
}]{%
gueta_can_2024}
\APACinsertmetastar {%
gueta_can_2024}%
\begin{APACrefauthors}%
Gueta, A.%
, Feder, A.%
, Gekhman, Z.%
, Goldstein, A.%
\BCBL {}\ \BBA {} Reichart, R.%
\end{APACrefauthors}%
\unskip\
\newblock
\APACrefYearMonthDay{2024}{{\APACmonth{06}}}{}.
\newblock
\APACrefbtitle {Can {LLMs} {Learn} {Macroeconomic} {Narratives} from {Social} {Media}?} {Can {LLMs} {Learn} {Macroeconomic} {Narratives} from {Social} {Media}?}
\newblock
\APACaddressPublisher{}{arXiv}.
\newblock
\begin{APACrefURL} [{2024-06-26}]\url{http://arxiv.org/abs/2406.12109} \end{APACrefURL}
\newblock
\APACrefnote{arXiv:2406.12109 [cs]}
\PrintBackRefs{\CurrentBib}

\bibitem [\protect \citeauthoryear {%
Hong%
\ \protect \BOthers {.}}{%
Hong%
\ \protect \BOthers {.}}{%
{\protect \APACyear {2020}}%
}]{%
HONG2020}
\APACinsertmetastar {%
HONG2020}%
\begin{APACrefauthors}%
Hong, T.%
, Pinson, P.%
, Wang, Y.%
, Weron, R.%
, Yang, D.%
\BCBL {}\ \BBA {} Zareipour, H.%
\end{APACrefauthors}%
\unskip\
\newblock
\APACrefYearMonthDay{2020}{}{}.
\newblock
{\BBOQ}\APACrefatitle {{Energy Forecasting: A Review and Outlook}} {{Energy Forecasting: A Review and Outlook}}.{\BBCQ}
\newblock
\APACjournalVolNumPages{IEEE Open Access Journal of Power and Energy}{7}{}{376-388}.
\newblock
\begin{APACrefDOI} \doi{10.1109/OAJPE.2020.3029979} \end{APACrefDOI}
\PrintBackRefs{\CurrentBib}

\bibitem [\protect \citeauthoryear {%
Huang%
\ \BBA {} Kaewunruen%
}{%
Huang%
\ \BBA {} Kaewunruen%
}{%
{\protect \APACyear {2023}}%
}]{%
13_HuangJH}
\APACinsertmetastar {%
13_HuangJH}%
\begin{APACrefauthors}%
Huang, J.%
\BCBT {}\ \BBA {} Kaewunruen, S.%
\end{APACrefauthors}%
\unskip\
\newblock
\APACrefYearMonthDay{2023}{}{}.
\newblock
{\BBOQ}\APACrefatitle {{Forecasting Energy Consumption of a Public Building Using Transformer and Support Vector Regression}} {{Forecasting Energy Consumption of a Public Building Using Transformer and Support Vector Regression}}.{\BBCQ}
\newblock
\APACjournalVolNumPages{Energies}{16}{2}{}.
\newblock
\begin{APACrefURL} \url{https://www.mdpi.com/1996-1073/16/2/966} \end{APACrefURL}
\newblock
\begin{APACrefDOI} \doi{10.3390/en16020966} \end{APACrefDOI}
\PrintBackRefs{\CurrentBib}

\bibitem [\protect \citeauthoryear {%
Hyndman%
\ \BBA {} Athanasopoulos%
}{%
Hyndman%
\ \BBA {} Athanasopoulos%
}{%
{\protect \APACyear {2018}}%
}]{%
Hyndman2018}
\APACinsertmetastar {%
Hyndman2018}%
\begin{APACrefauthors}%
Hyndman, R\BPBI J.%
\BCBT {}\ \BBA {} Athanasopoulos, G.%
\end{APACrefauthors}%
\unskip\
\newblock
\APACrefYear{2018}.
\newblock
\APACrefbtitle {{Forecasting: Principles and Practice}} {{Forecasting: Principles and Practice}}\ (\PrintOrdinal{2nd}\ \BEd).
\newblock
\APACaddressPublisher{}{OTexts}.
\newblock
\begin{APACrefURL} \url{https://otexts.com/fpp2/} \end{APACrefURL}
\PrintBackRefs{\CurrentBib}

\bibitem [\protect \citeauthoryear {%
Jiang%
\ \protect \BOthers {.}}{%
Jiang%
\ \protect \BOthers {.}}{%
{\protect \APACyear {2024}}%
}]{%
17_Jiang}
\APACinsertmetastar {%
17_Jiang}%
\begin{APACrefauthors}%
Jiang, W.%
, Liu, B.%
, Liang, Y.%
, Gao, H.%
, Lin, P.%
, Zhang, D.%
\BCBL {}\ \BBA {} Hu, G.%
\end{APACrefauthors}%
\unskip\
\newblock
\APACrefYearMonthDay{2024}{}{}.
\newblock
{\BBOQ}\APACrefatitle {Applicability analysis of transformer to wind speed forecasting by a novel deep learning framework with multiple atmospheric variables} {Applicability analysis of transformer to wind speed forecasting by a novel deep learning framework with multiple atmospheric variables}.{\BBCQ}
\newblock
\APACjournalVolNumPages{Applied Energy}{353}{}{122155}.
\newblock
\begin{APACrefURL} \url{https://www.sciencedirect.com/science/article/pii/S0306261923015192} \end{APACrefURL}
\newblock
\begin{APACrefDOI} \doi{https://doi.org/10.1016/j.apenergy.2023.122155} \end{APACrefDOI}
\PrintBackRefs{\CurrentBib}

\bibitem [\protect \citeauthoryear {%
Jin%
\ \protect \BOthers {.}}{%
Jin%
\ \protect \BOthers {.}}{%
{\protect \APACyear {2024}}%
}]{%
jin_time-llm_2024}
\APACinsertmetastar {%
jin_time-llm_2024}%
\begin{APACrefauthors}%
Jin, M.%
, Wang, S.%
, Ma, L.%
, Chu, Z.%
, Zhang, J\BPBI Y.%
, Shi, X.%
\BDBL {}Wen, Q.%
\end{APACrefauthors}%
\unskip\
\newblock
\APACrefYearMonthDay{2024}{{\APACmonth{01}}}{}.
\newblock
\APACrefbtitle {Time-{LLM}: {Time} {Series} {Forecasting} by {Reprogramming} {Large} {Language} {Models}.} {Time-{LLM}: {Time} {Series} {Forecasting} by {Reprogramming} {Large} {Language} {Models}.}
\newblock
\APACaddressPublisher{}{arXiv}.
\newblock
\begin{APACrefURL} [{2024-06-26}]\url{http://arxiv.org/abs/2310.01728} \end{APACrefURL}
\newblock
\APACrefnote{arXiv:2310.01728 [cs]}
\newblock
\begin{APACrefDOI} \doi{10.48550/arXiv.2310.01728} \end{APACrefDOI}
\PrintBackRefs{\CurrentBib}

\bibitem [\protect \citeauthoryear {%
Jędrzejewski%
, Lago%
, Marcjasz%
\BCBL {}\ \BBA {} Weron%
}{%
Jędrzejewski%
\ \protect \BOthers {.}}{%
{\protect \APACyear {2022}}%
}]{%
Jedrzejewski2022}
\APACinsertmetastar {%
Jedrzejewski2022}%
\begin{APACrefauthors}%
Jędrzejewski, A.%
, Lago, J.%
, Marcjasz, G.%
\BCBL {}\ \BBA {} Weron, R.%
\end{APACrefauthors}%
\unskip\
\newblock
\APACrefYearMonthDay{2022}{}{}.
\newblock
{\BBOQ}\APACrefatitle {{Electricity Price Forecasting: The Dawn of Machine Learning}} {{Electricity Price Forecasting: The Dawn of Machine Learning}}.{\BBCQ}
\newblock
\APACjournalVolNumPages{IEEE Power and Energy Magazine}{20}{3}{24-31}.
\newblock
\begin{APACrefDOI} \doi{10.1109/MPE.2022.3150809} \end{APACrefDOI}
\PrintBackRefs{\CurrentBib}

\bibitem [\protect \citeauthoryear {%
Lago%
, Marcjasz%
, {De Schutter}%
\BCBL {}\ \BBA {} Weron%
}{%
Lago%
\ \protect \BOthers {.}}{%
{\protect \APACyear {2021}}%
}]{%
LAGO2021}
\APACinsertmetastar {%
LAGO2021}%
\begin{APACrefauthors}%
Lago, J.%
, Marcjasz, G.%
, {De Schutter}, B.%
\BCBL {}\ \BBA {} Weron, R.%
\end{APACrefauthors}%
\unskip\
\newblock
\APACrefYearMonthDay{2021}{}{}.
\newblock
{\BBOQ}\APACrefatitle {Forecasting day-ahead electricity prices: A review of state-of-the-art algorithms, best practices and an open-access benchmark} {Forecasting day-ahead electricity prices: A review of state-of-the-art algorithms, best practices and an open-access benchmark}.{\BBCQ}
\newblock
\APACjournalVolNumPages{Applied Energy}{293}{}{116983}.
\newblock
\begin{APACrefURL} \url{https://www.sciencedirect.com/science/article/pii/S0306261921004529} \end{APACrefURL}
\newblock
\begin{APACrefDOI} \doi{https://doi.org/10.1016/j.apenergy.2021.116983} \end{APACrefDOI}
\PrintBackRefs{\CurrentBib}

\bibitem [\protect \citeauthoryear {%
Liu%
\ \protect \BOthers {.}}{%
Liu%
\ \protect \BOthers {.}}{%
{\protect \APACyear {2024}}%
}]{%
liu_timecma_2024}
\APACinsertmetastar {%
liu_timecma_2024}%
\begin{APACrefauthors}%
Liu, C.%
, Xu, Q.%
, Miao, H.%
, Yang, S.%
, Zhang, L.%
, Long, C.%
\BDBL {}Zhao, R.%
\end{APACrefauthors}%
\unskip\
\newblock
\APACrefYearMonthDay{2024}{{\APACmonth{06}}}{}.
\newblock
\APACrefbtitle {{TimeCMA}: {Towards} {LLM}-{Empowered} {Time} {Series} {Forecasting} via {Cross}-{Modality} {Alignment}.} {{TimeCMA}: {Towards} {LLM}-{Empowered} {Time} {Series} {Forecasting} via {Cross}-{Modality} {Alignment}.}
\newblock
\APACaddressPublisher{}{arXiv}.
\newblock
\begin{APACrefURL} [{2024-06-26}]\url{http://arxiv.org/abs/2406.01638} \end{APACrefURL}
\newblock
\APACrefnote{arXiv:2406.01638 [cs]}
\newblock
\begin{APACrefDOI} \doi{10.48550/arXiv.2406.01638} \end{APACrefDOI}
\PrintBackRefs{\CurrentBib}

\bibitem [\protect \citeauthoryear {%
López~Santos%
, García-Santiago%
, Echevarría~Camarero%
, Blázquez~Gil%
\BCBL {}\ \BBA {} Carrasco~Ortega%
}{%
López~Santos%
\ \protect \BOthers {.}}{%
{\protect \APACyear {2022}}%
}]{%
24_Santos}
\APACinsertmetastar {%
24_Santos}%
\begin{APACrefauthors}%
López~Santos, M.%
, García-Santiago, X.%
, Echevarría~Camarero, F.%
, Blázquez~Gil, G.%
\BCBL {}\ \BBA {} Carrasco~Ortega, P.%
\end{APACrefauthors}%
\unskip\
\newblock
\APACrefYearMonthDay{2022}{}{}.
\newblock
{\BBOQ}\APACrefatitle {{Application of Temporal Fusion Transformer for Day-Ahead PV Power Forecasting}} {{Application of Temporal Fusion Transformer for Day-Ahead PV Power Forecasting}}.{\BBCQ}
\newblock
\APACjournalVolNumPages{Energies}{15}{14}{}.
\newblock
\begin{APACrefURL} \url{https://www.mdpi.com/1996-1073/15/14/5232} \end{APACrefURL}
\newblock
\begin{APACrefDOI} \doi{10.3390/en15145232} \end{APACrefDOI}
\PrintBackRefs{\CurrentBib}

\bibitem [\protect \citeauthoryear {%
Meng%
\ \protect \BOthers {.}}{%
Meng%
\ \protect \BOthers {.}}{%
{\protect \APACyear {2024}}%
}]{%
meng2024enhancing}
\APACinsertmetastar {%
meng2024enhancing}%
\begin{APACrefauthors}%
Meng, S.%
, Chen, A.%
, Wang, C.%
, Zheng, M.%
, Wu, F.%
, Chen, X.%
\BDBL {}Li, P.%
\end{APACrefauthors}%
\unskip\
\newblock
\APACrefYearMonthDay{2024}{}{}.
\newblock
{\BBOQ}\APACrefatitle {{Enhancing Exchange Rate Forecasting with Explainable Deep Learning Models}} {{Enhancing Exchange Rate Forecasting with Explainable Deep Learning Models}}.{\BBCQ}
\newblock
\APACjournalVolNumPages{arXiv preprint arXiv:2410.19241}{}{}{}.
\newblock
\begin{APACrefURL} \url{https://arxiv.org/abs/2410.19241v1} \end{APACrefURL}
\PrintBackRefs{\CurrentBib}

\bibitem [\protect \citeauthoryear {%
Mirchandani%
\ \protect \BOthers {.}}{%
Mirchandani%
\ \protect \BOthers {.}}{%
{\protect \APACyear {2023}}%
}]{%
mirchandani2023largelanguagemodelsgeneral}
\APACinsertmetastar {%
mirchandani2023largelanguagemodelsgeneral}%
\begin{APACrefauthors}%
Mirchandani, S.%
, Xia, F.%
, Florence, P.%
, Ichter, B.%
, Driess, D.%
, Arenas, M\BPBI G.%
\BDBL {}Zeng, A.%
\end{APACrefauthors}%
\unskip\
\newblock
\APACrefYearMonthDay{2023}{}{}.
\newblock
\APACrefbtitle {{Large Language Models as General Pattern Machines}.} {{Large Language Models as General Pattern Machines}.}
\newblock
\begin{APACrefURL} \url{https://arxiv.org/abs/2307.04721} \end{APACrefURL}
\PrintBackRefs{\CurrentBib}

\bibitem [\protect \citeauthoryear {%
Ni%
, Yu%
, Liu%
, Li%
\BCBL {}\ \BBA {} Lin%
}{%
Ni%
\ \protect \BOthers {.}}{%
{\protect \APACyear {2023}}%
}]{%
Basisformer}
\APACinsertmetastar {%
Basisformer}%
\begin{APACrefauthors}%
Ni, Z.%
, Yu, H.%
, Liu, S.%
, Li, J.%
\BCBL {}\ \BBA {} Lin, W.%
\end{APACrefauthors}%
\unskip\
\newblock
\APACrefYearMonthDay{2023}{}{}.
\newblock
{\BBOQ}\APACrefatitle {{Basisformer: Attention-based time series forecasting with learnable and interpretable basis}} {{Basisformer: Attention-based time series forecasting with learnable and interpretable basis}}.{\BBCQ}
\newblock
\BIn{} \APACrefbtitle {{Neural Information Processing Systems}.} {{Neural Information Processing Systems}.}
\PrintBackRefs{\CurrentBib}

\bibitem [\protect \citeauthoryear {%
Nie%
, Kong%
\BCBL {}\ \protect \BOthers {.}}{%
Nie%
, Kong%
\BCBL {}\ \protect \BOthers {.}}{%
{\protect \APACyear {2024}}%
}]{%
nie_survey_2024}
\APACinsertmetastar {%
nie_survey_2024}%
\begin{APACrefauthors}%
Nie, Y.%
, Kong, Y.%
, Dong, X.%
, Mulvey, J\BPBI M.%
, Poor, H\BPBI V.%
, Wen, Q.%
\BCBL {}\ \BBA {} Zohren, S.%
\end{APACrefauthors}%
\unskip\
\newblock
\APACrefYearMonthDay{2024}{{\APACmonth{06}}}{}.
\newblock
\APACrefbtitle {A {Survey} of {Large} {Language} {Models} for {Financial} {Applications}: {Progress}, {Prospects} and {Challenges}.} {A {Survey} of {Large} {Language} {Models} for {Financial} {Applications}: {Progress}, {Prospects} and {Challenges}.}
\newblock
\APACaddressPublisher{}{arXiv}.
\newblock
\begin{APACrefURL} [{2024-06-26}]\url{http://arxiv.org/abs/2406.11903} \end{APACrefURL}
\newblock
\APACrefnote{arXiv:2406.11903 [cs, q-fin]}
\PrintBackRefs{\CurrentBib}

\bibitem [\protect \citeauthoryear {%
Nie%
, Nguyen%
, Sinthong%
\BCBL {}\ \BBA {} Kalagnanam%
}{%
Nie%
, Nguyen%
\BCBL {}\ \protect \BOthers {.}}{%
{\protect \APACyear {2024}}%
}]{%
PatchTST}
\APACinsertmetastar {%
PatchTST}%
\begin{APACrefauthors}%
Nie, Y.%
, Nguyen, N\BPBI H.%
, Sinthong, P.%
\BCBL {}\ \BBA {} Kalagnanam, J.%
\end{APACrefauthors}%
\unskip\
\newblock
\APACrefYearMonthDay{2024}{}{}.
\newblock
{\BBOQ}\APACrefatitle {{A time series is worth 64 words: Long-term forecasting with transformers.}} {{A time series is worth 64 words: Long-term forecasting with transformers.}}{\BBCQ}
\newblock
\BIn{} \APACrefbtitle {{International Conference on Learning Representations}.} {{International Conference on Learning Representations}.}
\PrintBackRefs{\CurrentBib}

\bibitem [\protect \citeauthoryear {%
Noguer~i Alonso%
\ \BBA {} Franklin%
}{%
Noguer~i Alonso%
\ \BBA {} Franklin%
}{%
{\protect \APACyear {2024}}%
}]{%
alonso2024large}
\APACinsertmetastar {%
alonso2024large}%
\begin{APACrefauthors}%
Noguer~i Alonso, M.%
\BCBT {}\ \BBA {} Franklin, R\BPBI P.%
\end{APACrefauthors}%
\unskip\
\newblock
\APACrefYearMonthDay{2024}{October 15}{}.
\newblock
{\BBOQ}\APACrefatitle {{Large Language Models for Financial Time Series Forecasting}} {{Large Language Models for Financial Time Series Forecasting}}.{\BBCQ}
\newblock
\APACjournalVolNumPages{SSRN}{}{}{}.
\newblock
\begin{APACrefURL} \url{https://papers.ssrn.com/sol3/papers.cfm?abstract_id=4988022} \end{APACrefURL}
\PrintBackRefs{\CurrentBib}

\bibitem [\protect \citeauthoryear {%
Nowotarski%
\ \BBA {} Weron%
}{%
Nowotarski%
\ \BBA {} Weron%
}{%
{\protect \APACyear {2018}}%
}]{%
NOWOTARSKI2018}
\APACinsertmetastar {%
NOWOTARSKI2018}%
\begin{APACrefauthors}%
Nowotarski, J.%
\BCBT {}\ \BBA {} Weron, R.%
\end{APACrefauthors}%
\unskip\
\newblock
\APACrefYearMonthDay{2018}{}{}.
\newblock
{\BBOQ}\APACrefatitle {Recent advances in electricity price forecasting: A review of probabilistic forecasting} {Recent advances in electricity price forecasting: A review of probabilistic forecasting}.{\BBCQ}
\newblock
\APACjournalVolNumPages{Renewable and Sustainable Energy Reviews}{81}{}{1548-1568}.
\newblock
\begin{APACrefURL} \url{https://www.sciencedirect.com/science/article/pii/S1364032117308808} \end{APACrefURL}
\newblock
\begin{APACrefDOI} \doi{https://doi.org/10.1016/j.rser.2017.05.234} \end{APACrefDOI}
\PrintBackRefs{\CurrentBib}

\bibitem [\protect \citeauthoryear {%
Oh%
, Oh%
, Shin%
, Um%
\BCBL {}\ \BBA {} Kim%
}{%
Oh%
\ \protect \BOthers {.}}{%
{\protect \APACyear {2023}}%
}]{%
56_Oh}
\APACinsertmetastar {%
56_Oh}%
\begin{APACrefauthors}%
Oh, S.%
, Oh, S.%
, Shin, H.%
, Um, T\BHBI w.%
\BCBL {}\ \BBA {} Kim, J.%
\end{APACrefauthors}%
\unskip\
\newblock
\APACrefYearMonthDay{2023}{}{}.
\newblock
{\BBOQ}\APACrefatitle {{Deep Learning Model Performance and Optimal Model Study for Hourly Fine Power Consumption Prediction}} {{Deep Learning Model Performance and Optimal Model Study for Hourly Fine Power Consumption Prediction}}.{\BBCQ}
\newblock
\APACjournalVolNumPages{Electronics}{12}{16}{}.
\newblock
\begin{APACrefURL} \url{https://www.mdpi.com/2079-9292/12/16/3528} \end{APACrefURL}
\newblock
\begin{APACrefDOI} \doi{10.3390/electronics12163528} \end{APACrefDOI}
\PrintBackRefs{\CurrentBib}

\bibitem [\protect \citeauthoryear {%
Ramos%
, Villela%
, Silva%
\BCBL {}\ \BBA {} Dias%
}{%
Ramos%
\ \protect \BOthers {.}}{%
{\protect \APACyear {2023}}%
}]{%
14_Ramos}
\APACinsertmetastar {%
14_Ramos}%
\begin{APACrefauthors}%
Ramos, P\BPBI V\BPBI B.%
, Villela, S\BPBI M.%
, Silva, W\BPBI N.%
\BCBL {}\ \BBA {} Dias, B\BPBI H.%
\end{APACrefauthors}%
\unskip\
\newblock
\APACrefYearMonthDay{2023}{}{}.
\newblock
{\BBOQ}\APACrefatitle {Residential energy consumption forecasting using deep learning models} {Residential energy consumption forecasting using deep learning models}.{\BBCQ}
\newblock
\APACjournalVolNumPages{Applied Energy}{350}{}{121705}.
\newblock
\begin{APACrefURL} \url{https://www.sciencedirect.com/science/article/pii/S0306261923010693} \end{APACrefURL}
\newblock
\begin{APACrefDOI} \doi{https://doi.org/10.1016/j.apenergy.2023.121705} \end{APACrefDOI}
\PrintBackRefs{\CurrentBib}

\bibitem [\protect \citeauthoryear {%
Rasul%
\ \protect \BOthers {.}}{%
Rasul%
\ \protect \BOthers {.}}{%
{\protect \APACyear {2024}}%
}]{%
rasul_lag-llama_2024-1}
\APACinsertmetastar {%
rasul_lag-llama_2024-1}%
\begin{APACrefauthors}%
Rasul, K.%
, Ashok, A.%
, Williams, A\BPBI R.%
, Ghonia, H.%
, Bhagwatkar, R.%
, Khorasani, A.%
\BDBL {}Rish, I.%
\end{APACrefauthors}%
\unskip\
\newblock
\APACrefYearMonthDay{2024}{{\APACmonth{02}}}{}.
\newblock
\APACrefbtitle {Lag-{Llama}: {Towards} {Foundation} {Models} for {Probabilistic} {Time} {Series} {Forecasting}.} {Lag-{Llama}: {Towards} {Foundation} {Models} for {Probabilistic} {Time} {Series} {Forecasting}.}
\newblock
\APACaddressPublisher{}{arXiv}.
\newblock
\begin{APACrefURL} [{2024-06-26}]\url{http://arxiv.org/abs/2310.08278} \end{APACrefURL}
\newblock
\APACrefnote{arXiv:2310.08278 [cs]}
\newblock
\begin{APACrefDOI} \doi{10.48550/arXiv.2310.08278} \end{APACrefDOI}
\PrintBackRefs{\CurrentBib}

\bibitem [\protect \citeauthoryear {%
Robinson%
, Arcos-Vargas%
, Tennican%
\BCBL {}\ \BBA {} Núñez%
}{%
Robinson%
\ \protect \BOthers {.}}{%
{\protect \APACyear {2023}}%
}]{%
Robinson2023}
\APACinsertmetastar {%
Robinson2023}%
\begin{APACrefauthors}%
Robinson, D.%
, Arcos-Vargas, A.%
, Tennican, M.%
\BCBL {}\ \BBA {} Núñez, F.%
\end{APACrefauthors}%
\unskip\
\newblock
\APACrefYearMonthDay{2023}{}{}.
\newblock
{\BBOQ}\APACrefatitle {{The Iberian Exception: An overview of its effects over its first 100 days}} {{The Iberian Exception: An overview of its effects over its first 100 days}}.{\BBCQ}
\newblock
\APACjournalVolNumPages{arXiv preprint arXiv:2309.02608}{}{}{}.
\newblock
\APACrefnote{Submitted on 5 Sep 2023 (v1), last revised 25 Sep 2023 (this version, v2)}
\PrintBackRefs{\CurrentBib}

\bibitem [\protect \citeauthoryear {%
Seifert%
\ \BBA {} Uhrig-Homburg%
}{%
Seifert%
\ \BBA {} Uhrig-Homburg%
}{%
{\protect \APACyear {2007}}%
}]{%
seifert2007modelling}
\APACinsertmetastar {%
seifert2007modelling}%
\begin{APACrefauthors}%
Seifert, J.%
\BCBT {}\ \BBA {} Uhrig-Homburg, M.%
\end{APACrefauthors}%
\unskip\
\newblock
\APACrefYearMonthDay{2007}{}{}.
\newblock
{\BBOQ}\APACrefatitle {Modelling jumps in electricity prices: theory and empirical evidence} {Modelling jumps in electricity prices: theory and empirical evidence}.{\BBCQ}
\newblock
\APACjournalVolNumPages{Review of Derivatives Research}{10}{}{59--85}.
\PrintBackRefs{\CurrentBib}

\bibitem [\protect \citeauthoryear {%
Sherozbek%
, Park%
, Akhtar%
\BCBL {}\ \BBA {} Yang%
}{%
Sherozbek%
\ \protect \BOthers {.}}{%
{\protect \APACyear {2023}}%
}]{%
11_Sherozbek}
\APACinsertmetastar {%
11_Sherozbek}%
\begin{APACrefauthors}%
Sherozbek, J.%
, Park, J.%
, Akhtar, M\BPBI S.%
\BCBL {}\ \BBA {} Yang, O\BHBI B.%
\end{APACrefauthors}%
\unskip\
\newblock
\APACrefYearMonthDay{2023}{}{}.
\newblock
{\BBOQ}\APACrefatitle {{Transformers-Based Encoder Model for Forecasting Hourly Power Output of Transparent Photovoltaic Module Systems}} {{Transformers-Based Encoder Model for Forecasting Hourly Power Output of Transparent Photovoltaic Module Systems}}.{\BBCQ}
\newblock
\APACjournalVolNumPages{Energies}{16}{3}{}.
\newblock
\begin{APACrefURL} \url{https://www.mdpi.com/1996-1073/16/3/1353} \end{APACrefURL}
\newblock
\begin{APACrefDOI} \doi{10.3390/en16031353} \end{APACrefDOI}
\PrintBackRefs{\CurrentBib}

\bibitem [\protect \citeauthoryear {%
Tschora%
, Pierre%
, Plantevit%
\BCBL {}\ \BBA {} Robardet%
}{%
Tschora%
\ \protect \BOthers {.}}{%
{\protect \APACyear {2022}}%
}]{%
tschora2022electricity}
\APACinsertmetastar {%
tschora2022electricity}%
\begin{APACrefauthors}%
Tschora, L.%
, Pierre, E.%
, Plantevit, M.%
\BCBL {}\ \BBA {} Robardet, C.%
\end{APACrefauthors}%
\unskip\
\newblock
\APACrefYearMonthDay{2022}{}{}.
\newblock
{\BBOQ}\APACrefatitle {Electricity price forecasting on the day-ahead market using machine learning} {Electricity price forecasting on the day-ahead market using machine learning}.{\BBCQ}
\newblock
\APACjournalVolNumPages{Applied Energy}{313}{}{118752}.
\newblock
\begin{APACrefURL} \url{https://doi.org/10.1016/j.apenergy.2022.118752} \end{APACrefURL}
\newblock
\begin{APACrefDOI} \doi{10.1016/j.apenergy.2022.118752} \end{APACrefDOI}
\PrintBackRefs{\CurrentBib}

\bibitem [\protect \citeauthoryear {%
Weron%
}{%
Weron%
}{%
{\protect \APACyear {2014}}%
}]{%
WERON2014}
\APACinsertmetastar {%
WERON2014}%
\begin{APACrefauthors}%
Weron, R.%
\end{APACrefauthors}%
\unskip\
\newblock
\APACrefYearMonthDay{2014}{}{}.
\newblock
{\BBOQ}\APACrefatitle {Electricity price forecasting: A review of the state-of-the-art with a look into the future} {Electricity price forecasting: A review of the state-of-the-art with a look into the future}.{\BBCQ}
\newblock
\APACjournalVolNumPages{International Journal of Forecasting}{30}{4}{1030-1081}.
\newblock
\begin{APACrefURL} \url{https://www.sciencedirect.com/science/article/pii/S0169207014001083} \end{APACrefURL}
\newblock
\begin{APACrefDOI} \doi{https://doi.org/10.1016/j.ijforecast.2014.08.008} \end{APACrefDOI}
\PrintBackRefs{\CurrentBib}

\bibitem [\protect \citeauthoryear {%
Wu%
, Xu%
, Wang%
\BCBL {}\ \BBA {} Long%
}{%
Wu%
\ \protect \BOthers {.}}{%
{\protect \APACyear {2021}}%
}]{%
Autoformer}
\APACinsertmetastar {%
Autoformer}%
\begin{APACrefauthors}%
Wu, H.%
, Xu, J.%
, Wang, J.%
\BCBL {}\ \BBA {} Long, M.%
\end{APACrefauthors}%
\unskip\
\newblock
\APACrefYearMonthDay{2021}{}{}.
\newblock
{\BBOQ}\APACrefatitle {{Autoformer: Decomposition transformers with auto-correlation for long-term series forecasting}} {{Autoformer: Decomposition transformers with auto-correlation for long-term series forecasting}}.{\BBCQ}
\newblock
\BIn{} \APACrefbtitle {{Advances Neural Information Processing Systems}.} {{Advances Neural Information Processing Systems}.}
\PrintBackRefs{\CurrentBib}

\bibitem [\protect \citeauthoryear {%
Xu%
\ \protect \BOthers {.}}{%
Xu%
\ \protect \BOthers {.}}{%
{\protect \APACyear {2023}}%
}]{%
3_Xu}
\APACinsertmetastar {%
3_Xu}%
\begin{APACrefauthors}%
Xu, J.%
, Hu, B.%
, Zhang, P.%
, Zhou, X.%
, Xing, Z.%
\BCBL {}\ \BBA {} Hu, Z.%
\end{APACrefauthors}%
\unskip\
\newblock
\APACrefYearMonthDay{2023}{}{}.
\newblock
{\BBOQ}\APACrefatitle {Regional electricity market price forecasting based on an adaptive spatial–temporal convolutional network} {Regional electricity market price forecasting based on an adaptive spatial–temporal convolutional network}.{\BBCQ}
\newblock
\APACjournalVolNumPages{Frontiers in Energy Research}{11}{}{}.
\newblock
\begin{APACrefURL} \url{https://www.frontiersin.org/articles/10.3389/fenrg.2023.1168944} \end{APACrefURL}
\newblock
\begin{APACrefDOI} \doi{10.3389/fenrg.2023.1168944} \end{APACrefDOI}
\PrintBackRefs{\CurrentBib}

\bibitem [\protect \citeauthoryear {%
Zeng%
, Chen%
, Zhang%
\BCBL {}\ \BBA {} Xu%
}{%
Zeng%
\ \protect \BOthers {.}}{%
{\protect \APACyear {2022}}%
}]{%
nlinear_dlinear}
\APACinsertmetastar {%
nlinear_dlinear}%
\begin{APACrefauthors}%
Zeng, A.%
, Chen, M.%
, Zhang, L.%
\BCBL {}\ \BBA {} Xu, Q.%
\end{APACrefauthors}%
\unskip\
\newblock
\APACrefYearMonthDay{2022}{}{}.
\newblock
\APACrefbtitle {{Are Transformers Effective for Time Series Forecasting?}} {{Are Transformers Effective for Time Series Forecasting?}}
\newblock
\begin{APACrefURL} \url{https://arxiv.org/abs/2205.13504} \end{APACrefURL}
\PrintBackRefs{\CurrentBib}

\bibitem [\protect \citeauthoryear {%
B.~Zhang%
, Song%
, Jiang%
\BCBL {}\ \BBA {} Li%
}{%
B.~Zhang%
\ \protect \BOthers {.}}{%
{\protect \APACyear {2023}}%
}]{%
2_Zhang}
\APACinsertmetastar {%
2_Zhang}%
\begin{APACrefauthors}%
Zhang, B.%
, Song, C.%
, Jiang, X.%
\BCBL {}\ \BBA {} Li, Y.%
\end{APACrefauthors}%
\unskip\
\newblock
\APACrefYearMonthDay{2023}{}{}.
\newblock
{\BBOQ}\APACrefatitle {{Electricity price forecast based on the STL-TCN-NBEATS model}} {{Electricity price forecast based on the STL-TCN-NBEATS model}}.{\BBCQ}
\newblock
\APACjournalVolNumPages{Heliyon}{9}{1}{e13029}.
\newblock
\begin{APACrefURL} \url{https://www.sciencedirect.com/science/article/pii/S2405844023002360} \end{APACrefURL}
\newblock
\begin{APACrefDOI} \doi{https://doi.org/10.1016/j.heliyon.2023.e13029} \end{APACrefDOI}
\PrintBackRefs{\CurrentBib}

\bibitem [\protect \citeauthoryear {%
X.~Zhang%
, Chowdhury%
, Gupta%
\BCBL {}\ \BBA {} Shang%
}{%
X.~Zhang%
\ \protect \BOthers {.}}{%
{\protect \APACyear {2024}}%
}]{%
zhang2024largelanguagemodelstime}
\APACinsertmetastar {%
zhang2024largelanguagemodelstime}%
\begin{APACrefauthors}%
Zhang, X.%
, Chowdhury, R\BPBI R.%
, Gupta, R\BPBI K.%
\BCBL {}\ \BBA {} Shang, J.%
\end{APACrefauthors}%
\unskip\
\newblock
\APACrefYearMonthDay{2024}{}{}.
\newblock
\APACrefbtitle {{Large Language Models for Time Series: A Survey}.} {{Large Language Models for Time Series: A Survey}.}
\newblock
\begin{APACrefURL} \url{https://arxiv.org/abs/2402.01801} \end{APACrefURL}
\PrintBackRefs{\CurrentBib}

\bibitem [\protect \citeauthoryear {%
Y.~Zhang%
\ \BBA {} Yan%
}{%
Y.~Zhang%
\ \BBA {} Yan%
}{%
{\protect \APACyear {2023}}%
}]{%
Crossformer}
\APACinsertmetastar {%
Crossformer}%
\begin{APACrefauthors}%
Zhang, Y.%
\BCBT {}\ \BBA {} Yan, J.%
\end{APACrefauthors}%
\unskip\
\newblock
\APACrefYearMonthDay{2023}{}{}.
\newblock
{\BBOQ}\APACrefatitle {{Crossformer: Transformer utilizing cross-dimension dependency for multivariate time series forecasting}} {{Crossformer: Transformer utilizing cross-dimension dependency for multivariate time series forecasting}}.{\BBCQ}
\newblock
\BIn{} \APACrefbtitle {{International Conference on Learning Representations}.} {{International Conference on Learning Representations}.}
\PrintBackRefs{\CurrentBib}

\bibitem [\protect \citeauthoryear {%
Zhao%
, Han%
, Ouyang%
\BCBL {}\ \BBA {} Burke%
}{%
Zhao%
\ \protect \BOthers {.}}{%
{\protect \APACyear {2023}}%
}]{%
62_Zhao}
\APACinsertmetastar {%
62_Zhao}%
\begin{APACrefauthors}%
Zhao, J.%
, Han, X.%
, Ouyang, M.%
\BCBL {}\ \BBA {} Burke, A\BPBI F.%
\end{APACrefauthors}%
\unskip\
\newblock
\APACrefYearMonthDay{2023}{}{}.
\newblock
{\BBOQ}\APACrefatitle {Specialized deep neural networks for battery health prognostics: Opportunities and challenges} {Specialized deep neural networks for battery health prognostics: Opportunities and challenges}.{\BBCQ}
\newblock
\APACjournalVolNumPages{Journal of Energy Chemistry}{87}{}{416-438}.
\newblock
\begin{APACrefURL} \url{https://www.sciencedirect.com/science/article/pii/S2095495623004977} \end{APACrefURL}
\newblock
\begin{APACrefDOI} \doi{https://doi.org/10.1016/j.jechem.2023.08.047} \end{APACrefDOI}
\PrintBackRefs{\CurrentBib}

\bibitem [\protect \citeauthoryear {%
Zhong%
}{%
Zhong%
}{%
{\protect \APACyear {2023}}%
}]{%
8_Zhong}
\APACinsertmetastar {%
8_Zhong}%
\begin{APACrefauthors}%
Zhong, B.%
\end{APACrefauthors}%
\unskip\
\newblock
\APACrefYearMonthDay{2023}{}{}.
\newblock
{\BBOQ}\APACrefatitle {{Deep learning integration optimization of electric energy load forecasting and market price based on the ANN–LSTM–transformer method}} {{Deep learning integration optimization of electric energy load forecasting and market price based on the ANN–LSTM–transformer method}}.{\BBCQ}
\newblock
\APACjournalVolNumPages{Frontiers in Energy Research}{11}{}{}.
\newblock
\begin{APACrefURL} \url{https://www.frontiersin.org/articles/10.3389/fenrg.2023.1292204} \end{APACrefURL}
\newblock
\begin{APACrefDOI} \doi{10.3389/fenrg.2023.1292204} \end{APACrefDOI}
\PrintBackRefs{\CurrentBib}

\bibitem [\protect \citeauthoryear {%
H.~Zhou%
\ \protect \BOthers {.}}{%
H.~Zhou%
\ \protect \BOthers {.}}{%
{\protect \APACyear {2021}}%
}]{%
Informer}
\APACinsertmetastar {%
Informer}%
\begin{APACrefauthors}%
Zhou, H.%
, Zhang, S.%
, Peng, J.%
, Zhang, S.%
, Li, J.%
, Xiong, H.%
\BCBL {}\ \BBA {} Zhang, W.%
\end{APACrefauthors}%
\unskip\
\newblock
\APACrefYearMonthDay{2021}{}{}.
\newblock
{\BBOQ}\APACrefatitle {{Informer: Beyond efficient transformer for long sequence time-series forecasting}} {{Informer: Beyond efficient transformer for long sequence time-series forecasting}}.{\BBCQ}
\newblock
\BIn{} \APACrefbtitle {{AAAI Conference on Artificial Intelligence}.} {{AAAI Conference on Artificial Intelligence}.}
\PrintBackRefs{\CurrentBib}

\bibitem [\protect \citeauthoryear {%
T.~Zhou%
\ \protect \BOthers {.}}{%
T.~Zhou%
\ \protect \BOthers {.}}{%
{\protect \APACyear {2022}}%
}]{%
FEDformer}
\APACinsertmetastar {%
FEDformer}%
\begin{APACrefauthors}%
Zhou, T.%
, Ma, Z.%
, Wen, Q.%
, Wang, X.%
, Sun, L.%
\BCBL {}\ \BBA {} Jin, R.%
\end{APACrefauthors}%
\unskip\
\newblock
\APACrefYearMonthDay{2022}{}{}.
\newblock
{\BBOQ}\APACrefatitle {{FEDformer: Frequency enhanced decomposed transformer for long-term series forecasting}} {{FEDformer: Frequency enhanced decomposed transformer for long-term series forecasting}}.{\BBCQ}
\newblock
\BIn{} \APACrefbtitle {{International Conference on Machine Learning}.} {{International Conference on Machine Learning}.}
\PrintBackRefs{\CurrentBib}

\bibitem [\protect \citeauthoryear {%
T.~Zhou%
, Niu%
, Wang%
, Sun%
\BCBL {}\ \BBA {} Jin%
}{%
T.~Zhou%
\ \protect \BOthers {.}}{%
{\protect \APACyear {2023}}%
}]{%
zhou_one_2023}
\APACinsertmetastar {%
zhou_one_2023}%
\begin{APACrefauthors}%
Zhou, T.%
, Niu, P.%
, Wang, X.%
, Sun, L.%
\BCBL {}\ \BBA {} Jin, R.%
\end{APACrefauthors}%
\unskip\
\newblock
\APACrefYearMonthDay{2023}{{\APACmonth{10}}}{}.
\newblock
\APACrefbtitle {One {Fits} {All}:{Power} {General} {Time} {Series} {Analysis} by {Pretrained} {LM}.} {One {Fits} {All}:{Power} {General} {Time} {Series} {Analysis} by {Pretrained} {LM}.}
\newblock
\APACaddressPublisher{}{arXiv}.
\newblock
\begin{APACrefURL} [{2024-06-26}]\url{http://arxiv.org/abs/2302.11939} \end{APACrefURL}
\newblock
\APACrefnote{arXiv:2302.11939 [cs]}
\newblock
\begin{APACrefDOI} \doi{10.48550/arXiv.2302.11939} \end{APACrefDOI}
\PrintBackRefs{\CurrentBib}

\end{thebibliography}

\newpage
\appendix
\appendixpage

\begin{appendices}
\section{Ranking of Time Series Models}\label{sec:Appendix1}
In this section, we provide an overview of the ranking of the models per country. In addition to that, we present the results from Friedman ranking test, which we utilize in order to verify that the results obtained from the best performing model are significantly different from all other models' predictions.\\

\begin{figure}[H]
\centering
\includegraphics[scale=0.85]{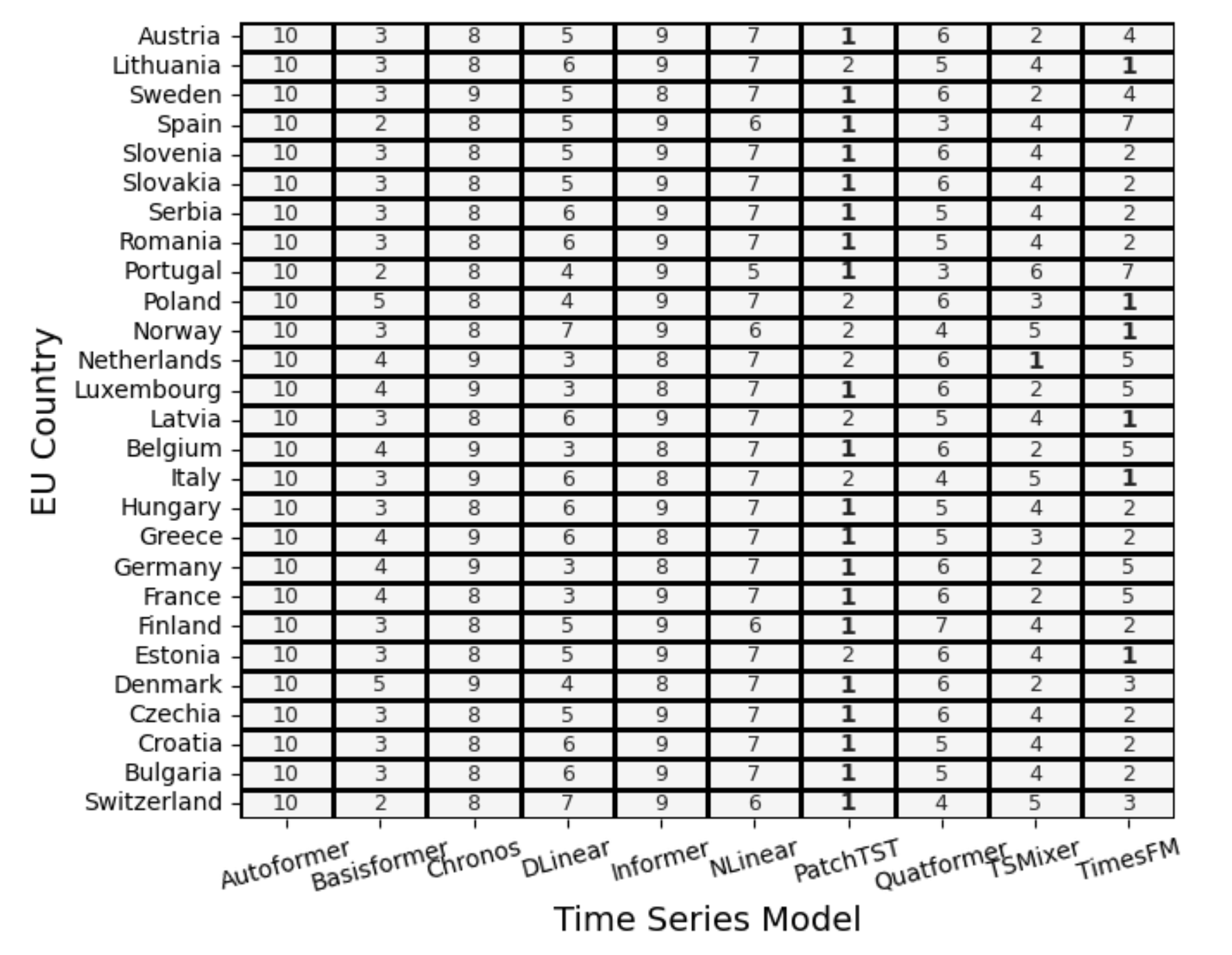}
\caption{Ranking Table}
\label{fig:rankingheatmap}
\end{figure}

The ranking table in Figure \ref{fig:rankingheatmap} reveals that PatchTST outperforms the majority of time series models, securing the best performance in 20 out of the 27 EU countries included in this research. 
TSMixer and TimesFM deliver the best performance on the countries, on which PatchTST does not produce the most accurate forecasts. In addition to that, Basisformer ranks in most of the cases on the second or the third spot.  
To further assess the performance differences, we applied the Friedman ranking test to evaluate the pairwise ranking differences between PatchTST and the other models. Table \ref{tab:pvalues} shows that the test yielded p-values of less than 0.05 for all comparisons, indicating that PatchTST’s results are statistically significantly different from those of the other models. 
\begin{table}[H]
\centering
\begin{tabular}{lll}
\toprule
\textbf{Model}      & \textbf{p-value}               & \textbf{Average rank}           \\
\midrule
TimesFM             & < 0.05                   & 2.926      \\
Basisformer         & < 0.00                   & 3.259      \\
TSMixer             & < 0.00                   & 3.481      \\
DLinear             & < 0.00                   & 5.000      \\
Quatformer          & < 0.00                   & 5.296      \\
NLinear             & < 0.00                   & 6.778      \\
Chronos             & < 0.00                   & 8.370      \\
Informer            & < 0.00                   & 8.852      \\
ARIMA               & < 0.00                   & 10.000     \\
Autoformer          & < 0.00                   & 10.778     \\
\bottomrule
\end{tabular}
\vspace{0.25cm}
\caption{P-values from Pairwise Friedman Ranking Test: comparison against PatchTST.}
\label{tab:pvalues}
\end{table}

\newpage
\section{Tabular Overview of Error Metrics for each Model per Country }
\label{sec:Appendix2}
\begin{table}[htbp]
    \centering
    \caption{Tables with the Achieved SMAPE, MAE, MSE, RMSE means across models and countries}
    \label{tab:my_label}
    \begin{subtable}[t]{0.45\textwidth}
        \centering
        \caption{ARIMA}
        \resizebox{\textwidth}{!}{%
        \begin{tabular}{lcccc}
        \toprule
        Country & SMAPE & RMSE & MSE & MAE \\
        \midrule
        Austria & 0.27 & 47.21 & 2569.37 & 43.08 \\
        Belgium & 0.22 & 32.31 & 1262.99 & 27.26 \\
        Bulgaria & 0.24 & 41.82 & 1973.36 & 36.14 \\
        Croatia & 0.46 & 48.40 & 2473.76 & 44.17 \\
        Czechia & 0.26 & 45.30 & 2384.86 & 40.87 \\
        Denmark & 0.25 & 40.75 & 1811.51 & 35.96 \\
        Estonia & 0.28 & 76.99 & 12984.28 & 60.11 \\
        Finland & 0.34 & 71.69 & 14762.11 & 53.67 \\
        France & 0.25 & 33.64 & 1379.25 & 28.87 \\
        Germany & 0.29 & 47.37 & 2605.04 & 42.83 \\
        Greece & 0.17 & 35.19 & 1341.70 & 29.03 \\
        Hungary & 0.23 & 39.48 & 1789.47 & 34.64 \\
        Italy & 0.18 & 42.12 & 1889.91 & 38.74 \\
        Latvia & 0.28 & 71.16 & 8441.47 & 57.58 \\
        Lithuania & 0.28 & 71.10 & 8437.47 & 57.55 \\
        Luxembourg & 0.29 & 47.37 & 2605.04 & 42.83 \\
        Netherlands & 0.83 & 65.60 & 4466.59 & 61.30 \\
        Norway & 0.17 & 27.43 & 858.50 & 23.19 \\
        Poland & 0.22 & 43.73 & 2186.16 & 40.14 \\
        Portugal & 0.42 & 50.42 & 3079.85 & 46.06 \\
        Romania & 0.25 & 43.84 & 2170.35 & 38.33 \\
        Serbia & 0.23 & 38.60 & 1698.76 & 34.18 \\
        Slovakia & 0.27 & 49.94 & 2816.19 & 45.41 \\
        Slovenia & 0.27 & 48.73 & 2707.89 & 44.43 \\
        Spain & 0.44 & 58.63 & 4001.79 & 53.92 \\
        Sweden & 0.25 & 39.20 & 1778.23 & 32.28 \\
        Switzerland & 0.28 & 53.51 & 3264.32 & 50.79 \\
        \bottomrule
        \end{tabular}
        }
    \end{subtable}
    \begin{subtable}[t]{0.45\textwidth}
        \centering
        \caption{Autoformer}
        \resizebox{\textwidth}{!}{%
        \begin{tabular}{lcccc}
        \toprule
        Country & SMAPE & RMSE & MSE & MAE \\
        \midrule
        Austria & 0.30 & 50.88 & 4360.95 & 46.15 \\
        Belgium & 0.27 & 36.70 & 1883.53 & 32.00 \\
        Bulgaria & 0.33 & 54.77 & 4848.79 & 49.23 \\
        Croatia & 0.34 & 49.42 & 3945.52 & 44.99 \\
        Czechia & 0.33 & 54.97 & 5866.72 & 50.80 \\
        Denmark & 0.29 & 58.36 & 8035.88 & 53.62 \\
        Estonia & 0.53 & 148.32 & 40828.85 & 132.57 \\
        Finland & 0.63 & 203.26 & 105368.34 & 187.68 \\
        France & 0.40 & 53.79 & 5234.79 & 49.63 \\
        Germany & 0.31 & 55.21 & 5300.56 & 50.33 \\
        Greece & 0.31 & 46.87 & 3834.26 & 41.97 \\
        Hungary & 0.44 & 68.21 & 7073.20 & 63.54 \\
        Italy & 0.13 & 29.94 & 1143.64 & 26.25 \\
        Latvia & 0.45 & 99.97 & 17616.94 & 86.27 \\
        Lithuania & 0.39 & 102.53 & 20233.02 & 88.38 \\
        Luxembourg & 0.31 & 51.20 & 4086.64 & 46.93 \\
        Netherlands & 0.36 & 47.39 & 2968.39 & 42.16 \\
        Norway & 0.21 & 38.08 & 2283.73 & 34.31 \\
        Poland & 0.31 & 65.36 & 8947.70 & 61.54 \\
        Portugal & 0.51 & 50.77 & 3394.96 & 45.81 \\
        Romania & 0.34 & 73.26 & 9848.97 & 67.93 \\
        Serbia & 0.25 & 43.95 & 3177.76 & 39.49 \\
        Slovakia & 0.27 & 40.52 & 2313.15 & 35.95 \\
        Slovenia & 0.31 & 49.19 & 3975.80 & 44.64 \\
        Spain & 0.48 & 44.87 & 2788.22 & 39.95 \\
        Sweden & 0.38 & 75.09 & 13103.26 & 68.55 \\
        Switzerland & 0.20 & 30.50 & 1264.19 & 26.68 \\
        \bottomrule
        \end{tabular}
        }
    \end{subtable}
    \begin{subtable}[t]{0.45\textwidth}
    \centering
    \caption{Basisformer}
    \resizebox{\textwidth}{!}{%
    \begin{tabular}{lcccc}
    \toprule
    Country & SMAPE & RMSE & MSE & MAE \\
    \midrule
    Austria & 0.18 & 22.13 & 610.36 & 18.03 \\
    Belgium & 0.21 & 23.56 & 680.46 & 19.38 \\
    Bulgaria & 0.18 & 24.74 & 707.68 & 19.61 \\
    Croatia & 0.17 & 22.04 & 596.98 & 17.86 \\
    Czechia & 0.18 & 22.34 & 621.44 & 18.01 \\
    Denmark & 0.19 & 26.56 & 855.89 & 21.81 \\
    Estonia & 0.19 & 57.71 & 11570.82 & 39.21 \\
    Finland & 0.27 & 61.82 & 12083.97 & 43.98 \\
    France & 0.22 & 23.23 & 661.08 & 19.24 \\
    Germany & 0.23 & 25.97 & 817.09 & 21.44 \\
    Greece & 0.12 & 22.17 & 544.65 & 16.93 \\
    Hungary & 0.17 & 23.34 & 652.20 & 18.74 \\
    Italy & 0.07 & 15.91 & 272.94 & 12.32 \\
    Latvia & 0.19 & 51.64 & 7129.45 & 36.30 \\
    Lithuania & 0.19 & 51.67 & 7131.53 & 36.32 \\
    Luxembourg & 0.23 & 25.97 & 817.09 & 21.44 \\
    Netherlands & 0.20 & 24.88 & 738.54 & 20.22 \\
    Norway & 0.12 & 20.14 & 585.59 & 15.91 \\
    Poland & 0.14 & 22.82 & 601.07 & 18.89 \\
    Portugal & 0.35 & 26.86 & 850.83 & 21.79 \\
    Romania & 0.18 & 24.93 & 721.59 & 19.81 \\
    Serbia & 0.15 & 21.43 & 555.09 & 17.48 \\
    Slovakia & 0.17 & 22.86 & 625.88 & 18.64 \\
    Slovenia & 0.17 & 21.94 & 595.10 & 17.85 \\
    Spain & 0.35 & 27.30 & 879.10 & 22.14 \\
    Sweden & 0.22 & 31.56 & 1381.75 & 25.17 \\
    Switzerland & 0.13 & 19.00 & 496.35 & 15.35 \\
    \bottomrule
    \end{tabular}
    }
\end{subtable}
\begin{subtable}[t]{0.45\textwidth}
    \centering
    \caption{Chronos}
    \resizebox{\textwidth}{!}{%
    \begin{tabular}{lcccc}
    \toprule
    Country & SMAPE & RMSE & MSE & MAE \\
    \midrule
    Austria & 0.25 & 28.25 & 1037.00 & 23.35 \\
    Belgium & 0.28 & 30.23 & 1129.26 & 24.97 \\
    Bulgaria & 0.24 & 30.44 & 1132.19 & 24.72 \\
    Croatia & 0.23 & 27.51 & 984.66 & 22.72 \\
    Czechia & 0.23 & 27.78 & 976.99 & 22.76 \\
    Denmark & 0.26 & 33.37 & 1381.34 & 27.45 \\
    Estonia & 0.23 & 62.75 & 12292.21 & 43.28 \\
    Finland & 0.34 & 68.19 & 12838.74 & 49.28 \\
    France & 0.29 & 28.93 & 1032.80 & 23.84 \\
    Germany & 0.31 & 32.87 & 1349.94 & 27.12 \\
    Greece & 0.16 & 27.70 & 942.35 & 22.04 \\
    Hungary & 0.22 & 28.18 & 999.62 & 23.03 \\
    Italy & 0.10 & 22.31 & 893.17 & 17.94 \\
    Latvia & 0.24 & 57.78 & 8079.41 & 41.35 \\
    Lithuania & 0.23 & 57.89 & 8584.79 & 41.72 \\
    Luxembourg & 0.31 & 32.99 & 1405.95 & 27.24 \\
    Netherlands & 0.27 & 31.89 & 1275.23 & 26.00 \\
    Norway & 0.15 & 24.01 & 854.27 & 19.04 \\
    Poland & 0.18 & 26.24 & 840.37 & 21.70 \\
    Portugal & 0.40 & 29.78 & 1130.64 & 24.11 \\
    Romania & 0.25 & 30.94 & 1174.49 & 25.28 \\
    Serbia & 0.19 & 25.04 & 794.57 & 20.78 \\
    Slovakia & 0.23 & 27.88 & 974.11 & 22.91 \\
    Slovenia & 0.23 & 27.06 & 944.75 & 22.33 \\
    Spain & 0.41 & 30.70 & 1177.19 & 24.80 \\
    Sweden & 0.30 & 37.29 & 1839.91 & 29.81 \\
    Switzerland & 0.17 & 24.31 & 837.62 & 19.82 \\
    \bottomrule
    \end{tabular}
    }
\end{subtable}
\end{table}

\newpage

\begin{table}[htbp]
    \centering
    \caption{Tables with the Achieved SMAPE, MAE, MSE, RMSE means across models and countries}
    \label{tab:my_label}
    \begin{subtable}[t]{0.45\textwidth}
    \centering
    \caption{DLinear}
    \resizebox{\textwidth}{!}{%
    \begin{tabular}{lcccc}
    \toprule
    Country & SMAPE & RMSE & MSE & MAE \\
    \midrule
    Austria & 0.17 & 24.32 & 751.44 & 19.67 \\
    Belgium & 0.19 & 24.80 & 753.78 & 19.93 \\
    Bulgaria & 0.18 & 27.81 & 930.05 & 22.25 \\
    Croatia & 0.17 & 24.89 & 782.92 & 20.16 \\
    Czechia & 0.17 & 24.80 & 768.53 & 19.96 \\
    Denmark & 0.18 & 27.47 & 854.65 & 21.82 \\
    Estonia & 0.21 & 61.76 & 11840.95 & 42.15 \\
    Finland & 0.27 & 64.34 & 12879.92 & 46.10 \\
    France & 0.20 & 24.68 & 747.06 & 20.10 \\
    Germany & 0.20 & 27.08 & 881.87 & 21.89 \\
    Greece & 0.13 & 24.22 & 631.76 & 18.57 \\
    Hungary & 0.17 & 26.27 & 863.81 & 21.26 \\
    Italy & 0.08 & 19.34 & 396.78 & 15.30 \\
    Latvia & 0.20 & 55.06 & 7152.48 & 38.69 \\
    Lithuania & 0.20 & 55.02 & 7149.47 & 38.64 \\
    Luxembourg & 0.20 & 27.08 & 881.87 & 21.89 \\
    Netherlands & 0.18 & 25.87 & 807.46 & 20.45 \\
    Norway & 0.13 & 22.91 & 820.04 & 17.92 \\
    Poland & 0.13 & 22.98 & 635.92 & 18.66 \\
    Portugal & 0.35 & 28.63 & 906.88 & 24.22 \\
    Romania & 0.18 & 28.11 & 957.92 & 22.56 \\
    Serbia & 0.15 & 23.94 & 724.25 & 19.60 \\
    Slovakia & 0.17 & 25.42 & 798.18 & 20.66 \\
    Slovenia & 0.17 & 24.50 & 763.61 & 19.87 \\
    Spain & 0.35 & 29.02 & 929.94 & 24.54 \\
    Sweden & 0.22 & 32.80 & 1565.58 & 25.79 \\
    Switzerland & 0.14 & 22.06 & 681.31 & 17.88 \\
    \bottomrule
    \end{tabular}%
    }
    \end{subtable}
    \begin{subtable}[t]{0.45\textwidth}
        \centering
        \caption{Informer}
        \resizebox{\textwidth}{!}{%
        \begin{tabular}{lcccc}
        \toprule
        Country & SMAPE & RMSE & MSE & MAE \\
        \midrule
        Austria & 0.22 & 33.42 & 1434.08 & 28.25 \\
    Belgium & 0.23 & 33.42 & 1428.57 & 28.24 \\
    Bulgaria & 0.23 & 39.07 & 1787.51 & 32.95 \\
    Croatia & 0.21 & 34.58 & 1469.60 & 29.29 \\
    Czechia & 0.22 & 34.73 & 1560.46 & 29.49 \\
    Denmark & 0.21 & 32.92 & 1285.18 & 27.07 \\
    Estonia & 0.27 & 78.40 & 13774.03 & 59.51 \\
    Finland & 0.31 & 79.00 & 14664.39 & 60.79 \\
    France & 0.25 & 34.08 & 1476.46 & 29.42 \\
    Germany & 0.23 & 32.88 & 1390.39 & 27.12 \\
    Greece & 0.15 & 29.16 & 887.71 & 22.57 \\
    Hungary & 0.22 & 35.72 & 1549.20 & 29.99 \\
    Italy & 0.09 & 22.59 & 529.55 & 18.22 \\
    Latvia & 0.25 & 66.68 & 8365.72 & 50.81 \\
    Lithuania & 0.25 & 66.01 & 8126.72 & 50.25 \\
    Luxembourg & 0.23 & 34.15 & 1522.68 & 28.51 \\
    Netherlands & 0.21 & 32.20 & 1298.54 & 26.01 \\
    Norway & 0.16 & 26.53 & 871.59 & 21.73 \\
    Poland & 0.16 & 29.68 & 1141.76 & 24.90 \\
    Portugal & 0.39 & 36.23 & 1539.88 & 31.66 \\
    Romania & 0.25 & 39.62 & 1874.86 & 33.43 \\
    Serbia & 0.21 & 34.76 & 1479.16 & 29.54 \\
    Slovakia & 0.21 & 33.72 & 1388.21 & 28.35 \\
    Slovenia & 0.21 & 33.49 & 1426.77 & 28.32 \\
    Spain & 0.39 & 36.24 & 1547.85 & 31.60 \\
    Sweden & 0.23 & 36.31 & 1683.63 & 28.94 \\
    Switzerland & 0.18 & 30.33 & 1281.89 & 25.89 \\
        \bottomrule
        \end{tabular}
        }
    \end{subtable}
    \begin{subtable}[t]{0.45\textwidth}
    \centering
    \caption{NLinear}
    \resizebox{\textwidth}{!}{%
    \begin{tabular}{lcccc}
    \toprule
    Country & SMAPE & RMSE & MSE & MAE \\
    \midrule
   Austria & 0.19 & 25.69 & 774.27 & 20.63 \\
    Belgium & 0.22 & 26.83 & 836.61 & 21.64 \\
    Bulgaria & 0.20 & 29.71 & 993.55 & 23.60 \\
    Croatia & 0.19 & 26.45 & 814.62 & 21.24 \\
    Czechia & 0.19 & 26.31 & 807.81 & 21.05 \\
    Denmark & 0.20 & 29.46 & 1019.51 & 23.58 \\
    Estonia & 0.21 & 63.26 & 12267.71 & 43.26 \\
    Finland & 0.28 & 65.02 & 12564.37 & 46.09 \\
    France & 0.23 & 26.13 & 795.06 & 21.20 \\
    Germany & 0.24 & 29.01 & 972.28 & 23.37 \\
    Greece & 0.13 & 25.74 & 720.93 & 19.73 \\
    Hungary & 0.19 & 28.20 & 918.26 & 22.49 \\
    Italy & 0.08 & 19.58 & 405.19 & 15.13 \\
    Latvia & 0.21 & 56.92 & 7656.42 & 40.26 \\
    Lithuania & 0.21 & 56.88 & 7653.36 & 40.23 \\
    Luxembourg & 0.24 & 29.01 & 972.28 & 23.37 \\
    Netherlands & 0.21 & 28.53 & 932.42 & 22.78 \\
    Norway & 0.12 & 22.12 & 682.32 & 17.12 \\
    Poland & 0.15 & 25.91 & 744.49 & 20.99 \\
    Portugal & 0.35 & 28.53 & 940.16 & 22.71 \\
    Romania & 0.20 & 29.88 & 1005.97 & 23.77 \\
    Serbia & 0.17 & 25.80 & 775.98 & 20.82 \\
    Slovakia & 0.19 & 27.07 & 839.70 & 21.80 \\
    Slovenia & 0.19 & 25.83 & 777.03 & 20.79 \\
    Spain & 0.36 & 29.08 & 974.09 & 23.15 \\
    Sweden & 0.22 & 33.92 & 1553.49 & 26.48 \\
    Switzerland & 0.14 & 21.51 & 587.23 & 17.11 \\
    \bottomrule
    \end{tabular}
    }
\end{subtable}
\begin{subtable}[t]{0.45\textwidth}
    \centering
    \caption{PatchTST}
    \resizebox{\textwidth}{!}{%
    \begin{tabular}{lcccc}
    \toprule
    Country & SMAPE & RMSE & MSE & MAE \\
    \midrule
   Austria & 0.16 & 20.36 & 529.82 & 16.51 \\
    Belgium & 0.19 & 22.18 & 615.17 & 18.12 \\
    Bulgaria & 0.16 & 22.69 & 612.97 & 17.78 \\
    Croatia & 0.15 & 19.93 & 504.20 & 16.03 \\
    Czechia & 0.17 & 20.94 & 553.78 & 16.79 \\
    Denmark & 0.18 & 25.13 & 778.42 & 20.37 \\
    Estonia & 0.18 & 56.54 & 11712.75 & 37.73 \\
    Finland & 0.26 & 60.98 & 12217.51 & 42.77 \\
    France & 0.21 & 21.92 & 600.95 & 18.00 \\
    Germany & 0.22 & 24.35 & 732.12 & 19.94 \\
    Greece & 0.11 & 20.08 & 440.66 & 15.20 \\
    Hungary & 0.16 & 21.33 & 559.48 & 16.92 \\
    Italy & 0.06 & 14.52 & 229.73 & 11.16 \\
    Latvia & 0.18 & 50.36 & 7117.71 & 34.80 \\
    Lithuania & 0.18 & 50.35 & 7116.88 & 34.79 \\
    Luxembourg & 0.22 & 24.35 & 732.12 & 19.94 \\
    Netherlands & 0.19 & 23.24 & 656.31 & 18.70 \\
    Norway & 0.11 & 19.10 & 530.56 & 14.80 \\
    Poland & 0.13 & 21.01 & 517.92 & 17.10 \\
    Portugal & 0.33 & 24.85 & 738.71 & 19.97 \\
    Romania & 0.17 & 22.85 & 625.37 & 17.95 \\
    Serbia & 0.13 & 18.89 & 448.72 & 15.34 \\
    Slovakia & 0.16 & 20.86 & 531.42 & 16.82 \\
    Slovenia & 0.16 & 20.01 & 510.10 & 16.17 \\
    Spain & 0.34 & 25.28 & 763.20 & 20.31 \\
    Sweden & 0.21 & 30.02 & 1272.94 & 23.52 \\
    Switzerland & 0.12 & 18.25 & 462.63 & 14.68 \\
    \bottomrule
    \end{tabular}
    }
\end{subtable}
\end{table}

\newpage

\begin{table}[htbp]
    \centering
    \caption{Tables with the Achieved SMAPE, MAE, MSE, RMSE means across models and countries}
    \label{tab:my_label}
    \begin{subtable}[t]{0.45\textwidth}
    \centering
    \caption{Quatformer}
    \resizebox{\textwidth}{!}{%
    \begin{tabular}{lcccc}
    \toprule
    Country & SMAPE & RMSE & MSE & MAE \\
    \midrule
    Austria & 0.19 & 23.36 & 668.93 & 19.26 \\
    Belgium & 0.22 & 24.48 & 725.90 & 20.30 \\
    Bulgaria & 0.18 & 25.68 & 762.54 & 20.45 \\
    Croatia & 0.18 & 23.06 & 647.21 & 18.83 \\
    Czechia & 0.19 & 23.38 & 674.26 & 19.09 \\
    Denmark & 0.20 & 27.42 & 894.18 & 22.73 \\
    Estonia & 0.23 & 59.54 & 11900.41 & 41.18 \\
    Finland & 0.29 & 64.51 & 12636.01 & 46.63 \\
    France & 0.23 & 24.43 & 718.57 & 20.34 \\
    Germany & 0.24 & 27.27 & 886.46 & 22.71 \\
    Greece & 0.13 & 23.69 & 612.26 & 18.46 \\
    Hungary & 0.17 & 23.92 & 678.65 & 19.39 \\
    Italy & 0.07 & 15.99 & 272.98 & 12.26 \\
    Latvia & 0.22 & 52.45 & 7221.50 & 37.44 \\
    Lithuania & 0.22 & 52.51 & 7265.44 & 37.52 \\
    Luxembourg & 0.24 & 27.37 & 893.73 & 22.80 \\
    Netherlands & 0.21 & 25.97 & 792.63 & 21.35 \\
    Norway & 0.12 & 20.71 & 588.65 & 16.53 \\
    Poland & 0.15 & 24.04 & 655.71 & 19.81 \\
    Portugal & 0.36 & 27.42 & 880.15 & 22.53 \\
    Romania & 0.19 & 25.79 & 769.45 & 20.66 \\
    Serbia & 0.17 & 22.54 & 604.88 & 18.54 \\
    Slovakia & 0.19 & 23.90 & 676.59 & 19.61 \\
    Slovenia & 0.19 & 23.04 & 647.39 & 18.94 \\
    Spain & 0.36 & 27.91 & 910.74 & 22.92 \\
    Sweden & 0.24 & 32.41 & 1409.55 & 26.14 \\
    Switzerland & 0.14 & 19.83 & 539.20 & 16.08 \\
    \bottomrule
    \end{tabular}%
    }
    \end{subtable}
    \begin{subtable}[t]{0.45\textwidth}
        \centering
        \caption{TSMixer}
        \resizebox{\textwidth}{!}{%
        \begin{tabular}{lcccc}
        \toprule
        Country & SMAPE & RMSE & MSE & MAE \\
        \midrule
          Austria & 0.17 & 22.83 & 683.64 & 18.61 \\
        Belgium & 0.18 & 23.62 & 699.10 & 19.19 \\
        Bulgaria & 0.17 & 25.79 & 809.14 & 20.52 \\
        Croatia & 0.16 & 23.34 & 708.12 & 19.02 \\
        Czechia & 0.17 & 23.54 & 718.96 & 19.04 \\
        Denmark & 0.18 & 26.15 & 777.95 & 20.89 \\
        Estonia & 0.19 & 59.03 & 11635.41 & 39.67 \\
        Finland & 0.26 & 63.05 & 12694.32 & 45.04 \\
        France & 0.20 & 23.99 & 720.40 & 19.90 \\
        Germany & 0.20 & 26.12 & 844.18 & 21.37 \\
        Greece & 0.12 & 22.24 & 542.98 & 16.83 \\
        Hungary & 0.16 & 24.59 & 779.52 & 19.96 \\
        Italy & 0.07 & 17.25 & 318.90 & 13.56 \\
        Latvia & 0.19 & 53.49 & 7079.08 & 37.38 \\
        Lithuania & 0.19 & 53.82 & 7099.26 & 37.75 \\
        Luxembourg & 0.20 & 25.94 & 820.84 & 21.18 \\
        Netherlands & 0.18 & 24.36 & 727.20 & 19.39 \\
        Norway & 0.12 & 21.48 & 679.25 & 16.85 \\
        Poland & 0.12 & 21.85 & 572.61 & 17.74 \\
        Portugal & 0.35 & 28.57 & 904.03 & 24.68 \\
        Romania & 0.17 & 26.18 & 845.04 & 20.96 \\
        Serbia & 0.15 & 22.22 & 628.45 & 18.16 \\
        Slovakia & 0.16 & 23.84 & 712.97 & 19.43 \\
        Slovenia & 0.16 & 23.01 & 694.90 & 18.78 \\
        Spain & 0.35 & 28.72 & 916.24 & 24.76 \\
        Sweden & 0.21 & 31.05 & 1409.74 & 24.38 \\
        Switzerland & 0.13 & 20.55 & 608.17 & 16.73 \\
        \bottomrule
        \end{tabular}
        }
    \end{subtable}
    \begin{subtable}[t]{0.45\textwidth}
    \centering
    \caption{TimesFM}
    \resizebox{\textwidth}{!}{%
    \begin{tabular}{lcccc}
    \toprule
    Country & SMAPE & RMSE & MSE & MAE \\
    \midrule
 Austria & 0.18 & 21.48 & 584.80 & 17.43 \\
        Belgium & 0.21 & 23.19 & 666.50 & 18.94 \\
        Bulgaria & 0.18 & 22.59 & 615.60 & 17.94 \\
        Croatia & 0.17 & 20.66 & 535.64 & 16.68 \\
        Czechia & 0.18 & 21.77 & 594.83 & 17.48 \\
        Denmark & 0.19 & 26.47 & 835.19 & 21.41 \\
        Estonia & 0.18 & 55.40 & 12005.78 & 36.62 \\
        Finland & 0.28 & 60.91 & 12574.44 & 42.88 \\
        France & 0.23 & 22.88 & 642.73 & 18.55 \\
        Germany & 0.24 & 26.14 & 832.18 & 21.43 \\
        Greece & 0.12 & 20.01 & 441.25 & 15.20 \\
        Hungary & 0.17 & 21.52 & 564.73 & 17.14 \\
        Italy & 0.06 & 14.15 & 224.20 & 10.74 \\
        Latvia & 0.18 & 49.09 & 7175.63 & 33.51 \\
        Lithuania & 0.18 & 48.95 & 7166.15 & 33.37 \\
        Luxembourg & 0.24 & 26.14 & 832.18 & 21.43 \\
        Netherlands & 0.21 & 24.46 & 726.06 & 19.74 \\
        Norway & 0.11 & 18.93 & 517.87 & 14.48 \\
        Poland & 0.13 & 20.60 & 506.12 & 16.69 \\
        Portugal & 0.40 & 25.07 & 793.69 & 19.89 \\
        Romania & 0.19 & 22.91 & 627.07 & 18.27 \\
        Serbia & 0.14 & 19.49 & 466.38 & 15.96 \\
        Slovakia & 0.17 & 21.57 & 565.45 & 17.43 \\
        Slovenia & 0.18 & 20.84 & 551.39 & 16.86 \\
        Spain & 0.41 & 25.58 & 820.22 & 20.24 \\
        Sweden & 0.23 & 31.12 & 1327.97 & 24.43 \\
        Switzerland & 0.13 & 19.58 & 543.50 & 15.53 \\
    \bottomrule
    \end{tabular}
    }
\end{subtable}
\end{table}

\end{appendices}
\end{document}